%% file: iclr2027_conference.tex
\documentclass{article} 
\usepackage{iclr2027_conference,times}
\iclrfinalcopy
\input{math_commands.tex}

\PassOptionsToPackage{numbers, compress}{natbib}
\usepackage{hyperref}
\usepackage{url}
\usepackage{amsmath}
\usepackage{graphicx}
\usepackage{multirow}
\usepackage{fancyvrb}
\usepackage{booktabs}

\usepackage{amssymb}

\usepackage{amsthm}

\theoremstyle{plain}
\newtheorem{theorem}{Theorem}
\newtheorem{lemma}[theorem]{Lemma}
\newtheorem{proposition}[theorem]{Proposition}
\newtheorem{corollary}[theorem]{Corollary}

\theoremstyle{definition}
\newtheorem{assumption}[theorem]{Assumption}

\title{TIPS: Topological Ill-Posedness Probing and Steering in Large Language Models}

\author{Guangyu Jiang \\
The George Washington University\\
\texttt{guangyu.jiang@gwu.edu} \\
\And
Sizhe Tang \\
The George Washington University\\
\texttt{s.tang1@gwu.edu} \\
\AND
Mahdi Imani \\
Northeastern University \\
\texttt{m.imani@northeastern.edu}
\AND
Nathaniel D. Bastian \\
Pathfinder Applied Research, LLC\\
\texttt{nbastian@pathfinderapplied}
\AND
Tian Lan \\
The George Washington University\\
\texttt{tlan@gwu.edu}
}
\begin{document}

\maketitle
\fancyhead{}
\lhead{Preprint}
\renewcommand{\headrulewidth}{0pt}

\begin{abstract}
Ill-posed questions, including those involving ambiguity, under-specification, or conflicting statements, may admit no valid answer or multiple plausible answers, posing a significant challenge for large language models (LLMs) despite their otherwise strong performance. Existing work often treats LLM reasoning as a black box and focuses on input-output analysis. Can a compact, unified topological representation of internal model states capture diverse sources of ill-posedness and steer reasoning toward responses appropriate to each source? To this end, we study the internal state from a topological perspective by treating the contextual hidden states of its prompt tokens at a single transformer layer as a point cloud, capturing the input problem’s relational structure as encoded by the model. Our analysis shows that zero- and one-dimensional persistent homology---tracking how connected components merge as the distance threshold increases (\(H_0\)) and the emergence and filling of one-dimensional loops in the filtration (\(H_1\)), respectively---yield seven compact descriptors of ill-posedness that also serve as control signals for steering model reasoning. Thus, we develop a topology-conditioned steering mechanism. It retrieves topologically similar positive and negative examples for each query to form a local activation-space contrast. The resulting query-specific direction modifies internal states during prefill and decoding, guiding the model's responses toward suitable abstention or clarification tailored to the underlying source of ill-posedness. Empirical evaluations on two datasets and open-weight models spanning four families demonstrate the effectiveness of this compact representation for ill-posedness classification, while topology-conditioning significantly improved steering performance.

\end{abstract}

\section{Introduction}
Large language models (LLMs) are increasingly deployed in open-world environments, where problems may be ill-posed---admitting no valid answer or multiple plausible answers~\citep{Hadamard1902}---rather than structured and well-defined. 
Many datasets and benchmarks show that ill-posed questions can originate from several distinct sources: ambiguities (lexical or semantic) with multiple plausible interpretations~\citep{min2020ambigqa, stelmakh2022asqa,zhang2024clamber}, underspecified entity or event references \citep{min2020ambigqa}, missing temporal or geographic context \citep{zhang2021situatedqa}, unclear user intent \citep{zhang2024clamber}, or over-specified problems with conflicting conditions \citep{ma2026large,xue2025reliablemath}. 
Models tend to falter once a question becomes ill-posed, where the difficulty lies as much in the question's structure as in the model's world knowledge.

Recent work shows that ill-posed problems can be broken into subclasses.
Notable advances include disambiguated input rewrites \citep{min2020ambigqa}, synthesizing answers across alternative interpretations \citep{stelmakh2022asqa}, selective abstention or rejection with respect to ambiguity \citep{cole2023selectively}, and active information acquisition and elicitation \citep{zhang2024clamber,aliannejadi2019asking, lee2023asking,fang2026mint}. These works establish that a strong response to ill-posed problems requires LLMs to adapt their behaviors to the underlying sources of ill-posedness. Crucially, such adaptation should not collapse to a fixed refusal template: a useful response should preserve the question-specific information, constraints, and context while directly addressing what is missing, ambiguous, or conflicting. However, most prior approaches treat LLM reasoning as a black box, addressing specific forms of ill-posedness through input-output analysis. Can a compact, unified topological representation of internal model states capture diverse sources of ill-posedness and steer reasoning toward responses appropriate to each source?


This paper demonstrates that the model's prefill stage provides a natural locus for such unified analysis and steering. Before generation begins, the model has already encoded the entire question into a sequence of layer-wise contextual token states. Recent work shows that intermediate activations can serve as both readout and intervention sites: to shift model behavior during inference~\citep{li2023inference}, enable contrastive activation steering~\citep{zou2023representation, rimsky2024steering}, and respond to unanswerability and abstention~\citep{arditi2024refusal,lavi2026detecting,zhang2025sparse,zhang2026locate}.    


We study this internal state from a topological perspective. Instead of representing a question by pooling hidden vectors, we treat the contextual hidden states of its prompt tokens at each transformer layer as a point cloud, capturing the input problem’s relational structure as encoded by the model.
We analyze this cloud using persistent homology,
which tracks topological features across a range of distance scales \citep{ghrist2008barcodes}. 
We show that zero- and one-dimensional persistent homology---tracking the merging of connected components as the distance threshold increases (\(H_0\)) and the emergence and filling of one-dimensional loops in the filtration (\(H_1\)), respectively---yield compact descriptors of ill-posedness that also serve as control signals for steering model reasoning. This perspective lets us quantify ill-posedness using
seven topological descriptors: three for \(H_0\) and four for \(H_1\). Under explicit class-dependent reference-geometry and
representation-distortion assumptions, we prove that mean finite \(H_0\) persistence and diameter-normalized maximum \(H_1\) persistence separate well-posed questions from nonunique classes with separated or cyclic reference structure (Theorem~\ref{thm:topology-separation}). This separation holds whenever the reference-class feature margins exceed the perturbation bounds induced by representation distortion.



These descriptors also serve as effective control signals for steering model reasoning in response to different forms of ill-posedness. We develop a topology-conditioned steering mechanism. Rather than applying a single global abstention direction that may produce generic refusals, our method retrieves topologically similar positive and negative examples for each query to form a local activation-space contrast.  The resulting query-specific direction modifies internal states during prefill and decoding, guiding the model responses toward suitable abstention or clarification tailored to the underlying source of ill-posedness. Persistent homology determines which samples enter the contrast for each case, while the intervention itself remains in the model's native activation space.

Evaluation on AmbigQA for binary ill-posedness detection, and on CLAMBER for fine-grained 9-way ill-posedness classification demonstrates the effectiveness of the proposed topological representation. 
Across eight open-weight models from the Llama, Qwen, Gemma, and Ministral families, our seven \(H_0\) and \(H_1\) descriptors from a
single selected layer improve average macro-F1 score over the strongest baselines from \(75.51\%\) to \(78.60\%\) on AmbigQA, and from \(36.64\%\) to \(39.44\%\) on CLAMBER 9-way classification. Averaged across the four dataset--model settings, topology-conditioned steering improves steered abstention response rate from \(27.43\%\) to \(51.15\%\). Together, these results demonstrate that compact token-cloud topology provides an effective, unified representation and steering signal for ill-posed problems.

Our contributions are threefold: 1) We introduce a layer-wise token-cloud representation for ill-posed question analysis, using persistent homology to capture the connectivity (\(H_0\)) and loop structure (\(H_1\)) of prompt-token hidden states.
2) We derive seven topological descriptors from a selected layer and evaluate binary and nine-way ill-posedness classification across eight models from four families and five split seeds. 3) We develop topology-conditioned activation steering using query-specific contrasts between topologically similar examples to encourage responses tailored to the source of ill-posedness.


\section{Related Work}

\paragraph{Ill-posedness, abstention, and clarification in question answering}
Prior work on ill-posed QA has studied several response behaviors. AmbigQA asks systems to recover multiple plausible question-answer pairs and produce disambiguated rewrites \citep{min2020ambigqa}, while ASQA requires long-form answers that synthesize information across different interpretations \citep{stelmakh2022asqa}. SituatedQA studies questions whose answers depend on temporal or geographic context \citep{zhang2021situatedqa}. Other work treats ill-posedness as a reason to abstain or clarify: \citet{cole2023selectively} distinguish uncertainty about the world from uncertainty about the question, while CLAMBER, Qulac, and CAmbigNQ evaluate clarification-question generation for underspecified information needs \citep{zhang2024clamber,aliannejadi2019asking, lee2023asking}. Alignment with Perceived Ambiguity further uses a model's own perceived ill-posedness signal to improve handling of unclear queries \citep{kim2024aligning}. These works primarily evaluate final response behavior, such as rewriting, answer synthesis, abstention, or clarification. Our work instead studies how ill-posedness is represented within the model before generation, using the layer-wise topology of question-token clouds for both detection- and abstention-oriented steering.

\paragraph{Internal representations of answerability, ill-posedness, and abstention}
A growing body of work suggests that LLMs encode reliability-relevant properties in intermediate activations. Inference-Time Intervention identifies truthfulness-related directions in attention-head activations \citep{li2023inference}, while representation engineering frames LLM control as monitoring and manipulating high-level internal representations \citep{zou2023representation}. Activation steering methods operationalize this idea by adding contrastive directions computed from positive and negative examples, as in Contrastive Activation Addition \citep{rimsky2024steering}. Related work shows that refusal can be mediated by low-dimensional activation directions \citep{arditi2024refusal}, and that unanswerability directions can be used for both detection and causal control of abstention \citep{lavi2026detecting}. Most relevant to our work, \citet{zhang2025sparse} show that question ambiguity, a central form of ill-posedness, is detectable during prefill and that a sparse set of neurons can shift responses from direct answering to abstention. Our framework complements these direction- and neuron-based approaches by representing each question as a layer-wise token cloud, covering fine-grained ill-posedness subtypes, and conditioning steering on topological similarity rather than a single global direction.

\paragraph{Topological data analysis for language representations}
Topological data analysis provides tools for summarizing the shape of high-dimensional point clouds. Persistent homology tracks connected components, cycles, and higher-dimensional holes across distance scales, producing diagrams or barcodes that can be vectorized for machine learning \citep{adams2017persistence}. Early NLP work used persistent homology as a text representation over embedded document units \citep{zhu2013persistent}. More recent work applies topology to language representation spaces, including cross-lingual word-embedding geometry \citep{draganov2024shape}, local topology of contextual language-model states \citep{ruppik2024local}, semantic-search ill-posedness over sentence-embedding neighborhoods \citep{barillot2024blowfish}, and zigzag-persistence analysis of layer-wise LLM dynamics \citep{gardinazzi2025persistent}. These works mainly use topology as a diagnostic representation tool. In contrast, we compute zero- and one-dimensional persistent homology over the token cloud of a single question at each prefill layer, relate the resulting descriptors to binary and fine-grained ill-posedness detection, and use topology to condition local activation steering from direct answering toward abstention.

\section{Method}
\label{sec:method}

\begin{figure}[t]
\centering
\includegraphics[width=\linewidth]{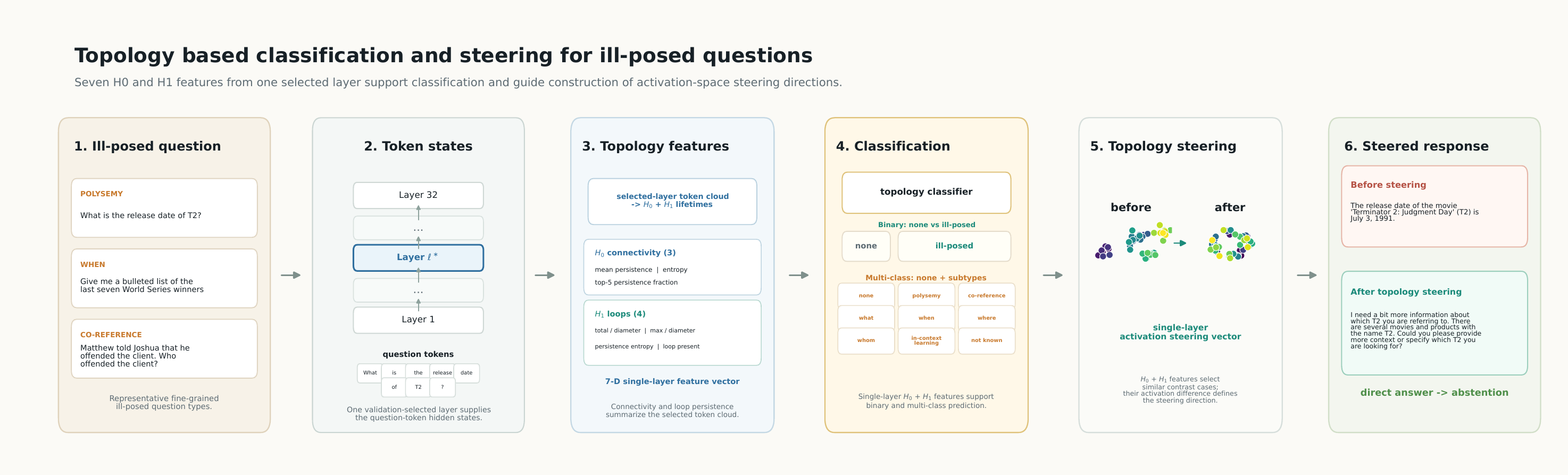}
\caption{Overview of our topology-based framework for ill-posedness classification and response steering. Prompt-token hidden states at a single selected transformer layer \(l^\star\) form a point cloud, summarized by seven descriptors of zero- and one-dimensional persistent homology. The resulting representation is used for binary and fine-grained classification and to retrieve topologically similar positive and negative examples. Their mean activation difference defines a query-specific
steering direction applied at the same layer to encourage source-aware clarification or abstention.}
\label{fig:method-overview}
\end{figure}

We propose a topology-based framework for analyzing ill-posed questions
and steering LLM responses. As shown in
Figure~\ref{fig:method-overview}, prompt-token hidden states at each
candidate transformer layer are treated as a point cloud and
characterized using zero- and one-dimensional persistent homology.
For downstream classification and steering, the seven descriptors
from a single selected layer form the question-level topological
representation.

\subsection{Layer-wise question token clouds}
\label{sec:token-clouds}
Let \(q\) be a question prompt containing \(T\) tokens and let \(M\) be a decoder-only language model with \(L\) transformer layers. During the prefill stage, the model processes all prompt tokens before generating the first output token. At layer \(l\), let \(h^{(l)}_t(q)\in\mathbb{R}^d\) denote the hidden state of the \(t\)-th prompt token, and let \(H^{(l)}(q)=[h^{(l)}_1(q),\ldots,h^{(l)}_T(q)]^\top\in\mathbb{R}^{T\times d}\) be the matrix obtained by stacking these \(T\) token states as rows. We view the rows of \(H^{(l)}(q)\) as a point cloud of contextual token representations. This representation preserves the token-level relational structure that mean pooling and last-token pooling discard. A pooled vector compresses the prompt into a single point, while the cloud retains the geometric arrangement of token states, which we measure with persistent homology. To reduce the cost of distance computation and control the representation dimension, we fit a layer-specific PCA map on prompt-token states from the training split and project each token cloud as \(\mathbf{X}^{(l)}(q)=[x^{(l)}_1(q),\ldots,x^{(l)}_T(q)]^\top=\left(H^{(l)}(q)-\mathbf{1}_T a_l^\top\right)P_l\in \mathbb{R}^{T\times r},\) where \(a_l\in\mathbb{R}^{d}\) is the mean over all training-split prompt-token states at layer \(l\), \(P_l\in\mathbb{R}^{d\times r}\) contains the top \(r\) principal directions, and \(\mathbf{1}_T\in\mathbb{R}^{T}\) is the all-ones vector. We denote the resulting point cloud in the reduced representation space by \(\mathcal{X}^{(l)}(q)=\{x^{(l)}_1(q),\ldots,x^{(l)}_T(q)\}\). 

\subsection{Persistent homology of question representations}
\label{sec:persistent-homology}

For each question \(q\) and candidate transformer layer \(l\), we compute persistent homology of the reduced token point cloud \(\mathcal{X}^{(l)}(q)=\{x_1^{(l)}(q),\ldots,x_T^{(l)}(q)\}\subset\mathbb{R}^{r}\) defined in Section~\ref{sec:token-clouds}. The resulting persistence diagrams characterize the geometry of contextual prompt-token states at a fixed prefill layer, rather than a pooled prompt representation or the generated response.

Given \(\mathcal{X}^{(l)}(q)\), we construct a Vietoris--Rips filtration \citep{vietoris1927hoheren}. For each scale parameter \(\epsilon\geq 0\), the Vietoris--Rips complex \(\mathrm{VR}_{\epsilon}(\mathcal{X}^{(l)}(q))\) contains a simplex \([x_{i_0}^{(l)}(q),\ldots,x_{i_m}^{(l)}(q)]\) whenever all pairwise distances among its vertices are at most \(\epsilon\):

\begin{equation}
\label{eq:vr-complex}
[x_{i_0}^{(l)}(q),\ldots,x_{i_m}^{(l)}(q)]\in\mathrm{VR}_{\epsilon}(\mathcal{X}^{(l)}(q))\quad\Longleftrightarrow\quad\left\|x_{i_a}^{(l)}(q)-x_{i_b}^{(l)}(q)\right\|_2\leq \epsilon\quad\text{for all }a,b.
\end{equation}

As \(\epsilon\) increases, the Vietoris--Rips complexes form a nested sequence, \(\mathrm{VR}_{\epsilon_1}\!\left(\mathcal X^{(l)}(q)\right)\subseteq\mathrm{VR}_{\epsilon_2}\!\left(\mathcal X^{(l)}(q)\right)\qquad\text{whenever}\epsilon_1\leq\epsilon_2.\) Persistent homology tracks the birth and death of topological features along this filtration. We compute homology in dimensions \(k\in\{0,1\}\) with coefficients modulo \(2\), i.e., over \(\mathbb F_2\). For each dimension \(k\), let \(D_k^{(l)}(q)=\left\{\left(b_{k,i}^{(l)}(q),d_{k,i}^{(l)}(q)\right)\right\}_{i=1}^{m_k^{(l)}(q)}\) denote the multiset of finite persistence intervals, where \(m_k^{(l)}(q)\) is the number of such intervals. For \(H_0\), we exclude the unique essential interval corresponding to the connected component that persists for the remainder of the filtration. The lifetime of interval \(i\) is \(\ell_{k,i}^{(l)}(q)=d_{k,i}^{(l)}(q)-b_{k,i}^{(l)}(q),\) and the corresponding lifetime multiset is \(\mathcal L_k^{(l)}(q)=\left\{\ell_{k,i}^{(l)}(q)\right\}_{i=1}^{m_k^{(l)}(q)}.\) A longer lifetime indicates that the corresponding topological feature persists over a wider range of distance scales.

\paragraph{Zero-dimensional persistence.}
Zero-dimensional persistent homology, \(H_0\), tracks connected components of the token cloud. At \(\epsilon=0\), each token point forms a separate component. As \(\epsilon\) increases, nearby components merge until a single connected component remains. For \(T\) distinct projected token states, excluding the essential bar leaves \(T-1\) finite \(H_0\) intervals. Since all \(H_0\) classes are born at zero, their finite lifetimes equal their death scales.

Under the edge-threshold convention in ~\eqref{eq:vr-complex}, the multiset of finite \(H_0\) death times equals the multiset of edge weights selected by Kruskal's algorithm for a minimum spanning tree, up to tie handling \citep{kleinberg2006algorithm}. Thus, the \(H_0\) diagram describes the multiscale component-merging structure of the token cloud, equivalently summarizing the distribution of its minimum-spanning-tree edge lengths.

\paragraph{One-dimensional persistence.}
One-dimensional persistent homology, \(H_1\), tracks loops in the token-cloud proximity structure. An \(H_1\) class is born when edges form a closed cycle that has not yet been filled by triangles, and it dies when subsequent simplices fill that cycle. The number of finite \(H_1\) bars varies across questions and layers and may be zero. Whereas \(H_0\) describes component mergers, \(H_1\) captures cyclic structure in the full Vietoris--Rips complex that is not recoverable from the minimum-spanning-tree edge-weight multiset.

Both dimensions are determined by the pairwise geometry of the token cloud, but they summarize complementary aspects of that geometry: connectivity for \(H_0\) and persistent cycles for \(H_1\). Their lifetime multisets provide the compact descriptors introduced next.

\subsection{Compact topological descriptors}
\label{sec:topological-descriptors}

For each question \(q\) and layer \(l\), we summarize the finite
lifetime multisets \(\mathcal L_0^{(l)}(q)\) and
\(\mathcal L_1^{(l)}(q)\) with seven scalar descriptors.
Suppressing the dependence on \(q\) and \(l\), write
\(m_k=|\mathcal L_k|\) and
\(S_k=\sum_{i=1}^{m_k}\ell_i^{(k)}\) for \(k\in\{0,1\}\),
and let \(\Delta=\max_{i,j}\|x_i-x_j\|_2\) denote the
token-cloud diameter.

For \(m_k\geq2\) and \(S_k>0\), normalized persistence entropy is \(E_k=-\frac{1}{\log m_k}\sum_{i=1}^{m_k}p_i^{(k)}\log p_i^{(k)},\qquad p_i^{(k)}=\frac{\ell_i^{(k)}}{S_k}.\) We set \(E_k=0\) when \(m_k\leq1\) or \(S_k=0\). 

\paragraph{\(H_0\) descriptors.}
We use mean finite persistence, normalized persistence entropy \(E_0\), and the top-five persistence fraction:
\begin{equation}
\label{eq:h0-descriptors}
\mu_0=\frac{S_0}{m_0},\qquad R_{0,5}=\frac{\sum_{j=1}^{\min(5,m_0)}\ell_{[j]}^{(0)}}{S_0},
\end{equation}
where \(\ell_{[1]}^{(0)}\geq\cdots\geq\ell_{[m_0]}^{(0)}\) are the finite \(H_0\) lifetimes sorted in decreasing order. If \(m_0=0\) or \(S_0=0\), we set \(\mu_0=R_{0,5}=0\). These descriptors capture the typical component-merging scale, the evenness of persistence across mergers, and the concentration of persistence in the largest merging events, respectively.

\paragraph{\(H_1\) descriptors.}
Let
\(
\ell_{\max}^{(1)}
=
\begin{cases}
\max_{1\leq i\leq m_1}\ell_i^{(1)}, & m_1>0,\\
0, & m_1=0.
\end{cases}
\)
We use diameter-normalized total and maximum persistence,
normalized persistence entropy \(E_1\), and a loop-existence
indicator:
\begin{equation}
\label{eq:h1-descriptors}
P_{1,\mathrm{tot}}
=
\frac{S_1}{\Delta},
\qquad
P_{1,\max}
=
\frac{\ell_{\max}^{(1)}}{\Delta},
\qquad
I_1
=
\mathbf{1}\{m_1>0\}.
\end{equation}
When \(m_1=0\), all four \(H_1\) descriptors are zero.
For a degenerate cloud with \(\Delta=0\), we set
\(P_{1,\mathrm{tot}}=P_{1,\max}=0\).
These descriptors capture aggregate and dominant loop persistence
relative to cloud size, the distribution of persistence across
loops, and the presence of finite positive-persistence \(H_1\)
features, respectively.

The layer-specific topology vector is \(\phi^{(l)}(q)=[\mu_0,E_0,R_{0,5},
P_{1,\mathrm{tot}},P_{1,\max},E_1,I_1]^\top
\in\mathbb R^7.\)

\subsection{Topological representation for ill-posedness classification}
\label{sec:topology-classification}

For each question \(q\), we use the topology descriptor from a selected transformer layer \(l^\star\) as the question-level representation: \(z(q)=\phi^{(l^\star)}(q)\in\mathbb{R}^{7}.\) The same layer \(l^\star\) is used for both ill-posedness classification and topology-conditioned activation steering. We standardize \(z(q)\) using coordinate-wise means and standard deviations computed on the training split to obtain \(\widetilde{z}(q)\). We then train a supervised classifier \(g_\theta:\mathbb R^7\rightarrow\mathcal P_C,\ \mathcal P_C=\left\{p\in\mathbb R_{\geq0}^C:\sum_{c=1}^{C}p_c=1\right\},\) where \(C=2\) for binary well-posed versus ill-posed detection and \(C\) is the number of classes for fine-grained ill-posedness classification.

\subsection{Topology-conditioned local steering}
\label{sec:topology-steering}

We use the selected-layer topology representation not only for classification, but also to construct query-specific steering directions. Contrastive activation steering commonly constructs a global direction from the activation difference between positive and negative examples~\citep{rimsky2024steering}. Such a global direction may be too coarse for ill-posed questions involving different entities, times, locations, task constraints, or interpretations. We instead retrieve behaviorally matched examples in topology space and construct a local activation-space direction.

Let \(\mathcal{D}^{+}\) denote training questions whose base responses are ill-posedness-aware, and let \(\mathcal{D}^{-}\) denote training questions whose base responses directly answer the ill-posed question. For a test question \(q\), we compute its standardized topology vector \(\widetilde{z}(q)\) and retrieve its \(k\) nearest neighbors \(\mathcal{N}_{k}^{+}(q)\subset\mathcal{D}^{+},\mathcal{N}_{k}^{-}(q)\subset\mathcal{D}^{-},\) using Euclidean distance in standardized topology space.

For each retrieved example \(q_i\), let \(\bar{h}^{(l^\star)}(q_i)=\frac{1}{T_i}\sum_{t=1}^{T_i}h_t^{(l^\star)}(q_i)\in\mathbb{R}^{d}\) denote its mean-pooled prompt activation at the same selected layer \(l^\star\). We define the local steering vector as
\begin{equation}
\label{eq:steer-direction-vector}
v_{\mathrm{local}}(q)=\frac{1}{|\mathcal{N}_{k}^{+}(q)|}\sum_{q_i\in\mathcal{N}_{k}^{+}(q)}\bar{h}^{(l^\star)}(q_i)-\frac{1}{|\mathcal{N}_{k}^{-}(q)|}\sum_{q_j\in\mathcal{N}_{k}^{-}(q)}\bar{h}^{(l^\star)}(q_j).
\end{equation}

We then add this vector to the hidden states at layer \(l^\star\). During prefill,
\begin{equation}
\label{eq:steering}
h_t^{(l^\star)}(q)\leftarrow h_t^{(l^\star)}(q)+\alpha v_{\mathrm{local}}(q),\qquad t=1,\ldots,T,
\end{equation}
and during decoding we apply the same update to the current generated-token hidden state at layer \(l^\star\). The scalar \(\alpha\) controls the steering strength. The topology representation is used to retrieve local contrast examples, whereas the steering direction is constructed and applied in the model's original activation space.

\subsection{Theoretical basis for identification}
\label{sec:theory}

We establish conditions under which the proposed descriptors distinguish well-posed from nonunique questions. Full assumptions and proofs are provided in Appendix~\ref{app:theory}.

\paragraph{Geometric distinction.}
Motivated by the concentration of approximate solutions in well-posed minimization~\citep{furi1970characterization}, we consider three semantic reference classes: regular well-posed questions \(\mathcal W\), nonunique questions with separated resolution families \(\mathcal I_0\), and nonunique questions with persistent cyclic structure \(\mathcal I_1\). Under Assumption~\ref{assump:class-geometry}, reference configurations in \(\mathcal W\) have mean \(H_0\) persistence at most \(a_{\mathrm W}\) and diameter-normalized maximum \(H_1\) persistence at most \(\omega\). The corresponding lower bounds are \(b_0\) for \(\mathcal I_0\) and \(\gamma\) for \(\mathcal I_1\), as derived in Appendix~\ref{app:separation}.

\begin{theorem}[Robust class separation]
\label{thm:topology-separation}
Suppose the observed and reference token configurations satisfy Assumption~\ref{assump:representation-distortion}: their indexed pairwise distances differ by at most \(\delta\), and their diameters are at least \(\Delta_{\min}>0\). Under the reference-class conditions above, if
\begin{equation}
\label{eq:main-separation-margins}
b_0-a_{\mathrm W}>2\delta,
\qquad
\gamma-\omega>\frac{6\delta}{\Delta_{\min}},
\end{equation}
then the rule
\begin{equation}
\label{eq:main-separation-rule}
\widehat Y(q)
=
\mathbf{1}\!\left\{
\mu_0(q)>\frac{a_{\mathrm W}+b_0}{2}
\ \text{or}\
P_{1,\max}(q)>\frac{\omega+\gamma}{2}
\right\}
\end{equation}
assigns label \(0\) to every \(q\in\mathcal W\) and label \(1\)
to every \(q\in\mathcal I_0\cup\mathcal I_1\).
\end{theorem}

The result follows by bounding the perturbations of mean \(H_0\) persistence and normalized maximum \(H_1\) persistence by
\(\delta\) and \(3\delta/\Delta_{\min}\), respectively, using component-merging stability and Rips-persistence
stability~\citep{chazal2014persistence}. Thus, reference-class differences remain distinguishable when they exceed representation distortion. The theorem establishes separation in two coordinates of the seven-dimensional descriptor for the specified geometric classes.

\section{Experiment}
\label{sec:experiments}
Our experiments evaluate three questions. First, do well-posed questions and different ill-posedness subtypes exhibit distinct distributions in the single-layer topology space? Second, can compact finite \(H_0\) and \(H_1\) descriptors detect binary ill-posedness and fine-grained ill-posedness subtypes? Third, can topology-conditioned neighborhoods improve activation steering toward ill-posedness-aware responses, including abstention and clarification?

\subsection{Datasets and Models}

We evaluate two ill-posedness-oriented QA benchmarks. 
For binary ill-posedness detection, we use AmbigQA, which studies open-domain questions that may admit multiple plausible interpretations and provides disambiguated rewrites for resolving them \citep{min2020ambigqa}. 
For fine-grained ill-posedness classification, we use CLAMBER as a 9-way task: one well-posed \textsc{None} class plus eight ill-posedness subtypes. These subtypes cover epistemic misalignment, including unfamiliar or missing-knowledge queries and contradictions induced by in-context examples; linguistic indeterminacy, including lexical polysemy and semantic or co-reference uncertainty; and missing output constraints, including unspecified personal, temporal, spatial, or task-specific information \citep{zhang2024clamber}. We extract internal representation of the questions from eight open-weight models spanning four families, with model sizes ranging from 8B to 70B: Llama 3.1 and 3.3~\citep{grattafiori2024llama,meta2024llama33}, Qwen3.5 and Qwen3.6~\citep{qwen2026qwen35,qwen2026qwen36}, Gemma 4~\citep{team2026gemma}, and Ministral 3~\citep{liu2026ministral}. 

\subsection{Class structure in topology space}
\label{sec:topology-visualization}

We first examine the class structure captured by the seven-dimensional topological representation. Using Llama-3.1-8B-Instruct, we extract the seven \(H_0\) and \(H_1\) descriptors from selected layers for questions in AmbigQA and CLAMBER. For visualization, we project these question-level feature vectors onto the first two principal components using PCA. Each point represents one question, with color indicating its class label. 

\begin{figure}[t]
    \centering
    \includegraphics[width=\linewidth]
    {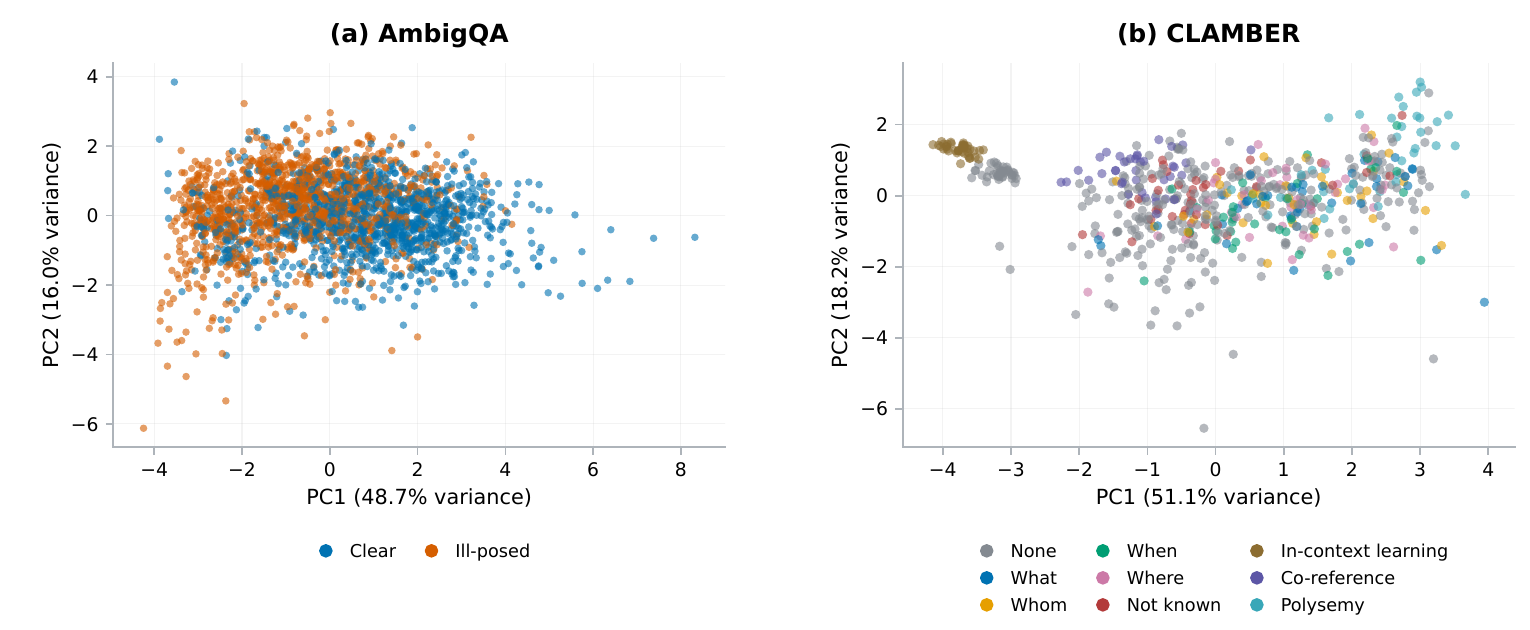}
    \caption{Two-dimensional PCA projections of the seven \(H_0\) and \(H_1\) descriptors at selected layer from Llama-3.1-8B-Instruct on (a) AmbigQA and (b) CLAMBER. Each point represents a question; colors indicate the binary labels in AmbigQA and the nine class labels in CLAMBER. The first two principal components explain \(64.7\%\) and \(69.3\%\) of the feature variance in the respective panels. The projections illustrate class-dependent structure with overlapping distributions.}
    \label{fig:topology-pca}
\end{figure}

Figure~\ref{fig:topology-pca} shows class-dependent organization in the projected feature space. On AmbigQA, clear and ill-posed
questions exhibit a partial distributional shift along the first principal component, although the two groups overlap substantially. On CLAMBER, the in-context-learning category forms a compact cluster at negative PC1 values, while several other subtypes overlap in the central region. These patterns provide a qualitative view of the distinctions captured by the topology descriptors and motivate the quantitative classification evaluation below.

\subsection{Ill-posedness classification}
\label{sec:classification}

We evaluate whether the seven descriptors of zero- and one-dimensional persistent homology support binary ill-posedness detection on AmbigQA~\citep{min2020ambigqa} and fine-grained nine-way classification on CLAMBER~\citep{zhang2024clamber}. For each question, we construct a seven-dimensional topology vector from a single transformer layer selected using validation-set classification performance. We train a logistic regression classifier on the standardized representation \(\widetilde z(q)\), with feature standardization fitted on the training split.

We compare against AEN \citep{zhang2025sparse}, a probing baseline that selects ambiguity-predictive neurons and trains a classifier on their activations. We use \(k=7\) neurons, to make its dimensionality to align with our topological features. Prompt-based comparisons include CLAMBER zero-shot~\citep{zhang2024clamber}, CLAM few-shot~\citep{kuhn2022clam}, CLAMBER few-shot chain-of-thought \citep{zhang2024clamber,wei2022chain}, and InfoGain prompting~\citep{kim2024aligning}.

We evaluate eight models from four families using five split seeds (0--4), with 70/10/20 train/validation/test splits. Layer selection is performed separately for each split using validation data. Table~\ref{tab:classification-results} reports macro-F1 as mean \(\pm\) sample standard deviation across the five splits.

\begin{table}[t]
\centering
\caption{
Ill-posedness classification: macro-F1 (mean \(\pm\) sample SD, \%) across five split seeds (0--4), using 70/10/20 train/validation/test splits. Ours uses seven \(H_0\) and \(H_1\) descriptors from one validation-selected layer; AEN uses seven neurons. InfoGain only works for the binary tasks and is therefore skipped for CLAMBER. Boldface marks the best method in each dataset--model row. Bold marks the highest available mean in each row.
}
\label{tab:classification-results}

\footnotesize
\setlength{\tabcolsep}{2pt}
\renewcommand{\arraystretch}{1.05}

\begin{tabular*}{\linewidth}{@{}l@{\extracolsep{\fill}}cccccc@{}}
\toprule
Model
& Ours
& AEN
& Zero-shot
& Few-shot
& Few-shot CoT
& InfoGain \\
\midrule

\multicolumn{7}{@{}l}{\textbf{AmbigQA (binary)}} \\

Llama-3.1-8B & $\mathbf{78.66}\pm0.78$ & $73.76\pm1.71$ & $45.15\pm0.68$ & $48.67\pm1.03$ & $40.69\pm0.82$ & $36.41\pm1.48$ \\

Llama-3.3-70B & $78.74\pm0.51$ & $\mathbf{79.89}\pm0.27$ & $33.58\pm0.83$ & $57.44\pm4.85$ & $35.60\pm1.94$ & $52.77\pm3.19$ \\

Qwen3.5-9B & $\mathbf{79.32}\pm1.38$ & $75.05\pm1.14$ & $43.12\pm0.61$ & $54.89\pm0.39$ & $37.10\pm0.21$ & $55.70\pm1.42$ \\

Qwen3.6-27B & $\mathbf{78.85}\pm0.87$ & $75.51\pm4.53$ &$47.12\pm1.20$ & $49.25\pm0.70$ & $48.05\pm0.37$ & $53.63\pm0.73$ \\

Gemma-4-12B & $\mathbf{78.35}\pm0.76$ & $68.09\pm2.99$ & $49.74\pm0.70$ & $62.10\pm0.94$ & $53.31\pm0.63$ & $52.55\pm0.42$ \\

Gemma-4-31B & $\mathbf{78.85}\pm0.50$ & $66.54\pm3.53$ & $49.46\pm0.32$ & $56.31\pm0.66$  & $52.33\pm0.82$ & $49.75\pm5.38$ \\

Ministral-3-8B & $78.01\pm0.55$ & $\mathbf{81.60}\pm0.69$ & $48.78\pm0.55$ & $48.50\pm0.79$ & $47.26\pm0.90$ & $36.12\pm2.65$ \\

Ministral-3-14B & $77.98\pm0.54$ & $\mathbf{83.63}\pm0.56$ & $54.78\pm1.11$ & $55.29\pm0.95$ & $49.43\pm0.34$ & $41.60\pm6.20$ \\

\midrule
\multicolumn{7}{@{}l}{\textbf{CLAMBER (9-way)}} \\

Llama-3.1-8B
& $\mathbf{40.72}\pm2.01$ & $34.53\pm1.18$ & $13.81\pm0.52$ & $26.83\pm1.80$ & $26.44\pm0.80$ & -- \\

Llama-3.3-70B
& $\mathbf{41.97}\pm1.76$ & $38.58\pm4.47$ & $27.82\pm4.30$  & $35.24\pm4.61$  & $33.76\pm2.71$ & -- \\

Qwen3.5-9B
& $\mathbf{41.00}\pm2.60$ & $37.11\pm6.73$ & $32.46\pm1.14$ & $33.68\pm0.84$ & $35.79\pm2.31$ & -- \\

Qwen3.6-27B
& $41.88\pm1.05$ & $36.24\pm3.43$ & $35.98\pm1.77$ & $\mathbf{42.12}\pm1.79$ &$40.03\pm1.61$ & -- \\

Gemma-4-12B
& $37.10\pm0.92$ & $35.44\pm3.61$ & $\mathbf{44.22}\pm2.16$ & $42.16\pm1.66$ & $43.96\pm1.00$ & -- \\

Gemma-4-31B
& $37.42\pm1.88$ & $34.07\pm4.30$ & $44.46\pm2.07$ & $44.03\pm2.45$ & $\mathbf{45.25}\pm2.02$ & -- \\

Ministral-3-8B
& $38.07\pm0.87$ & $\mathbf{41.81}\pm3.56$ & $26.99\pm1.66$ & $30.28\pm1.03$ & $25.90\pm1.45$ & -- \\

Ministral-3-14B
& $\mathbf{37.37}\pm1.84$ & $35.36\pm5.34$ & $32.20\pm1.95$ & $32.17\pm0.63$ & $19.65\pm1.16$ & -- \\

\bottomrule
\end{tabular*}
\end{table}

Table~\ref{tab:classification-results} shows that the seven selected-layer \(H_0\) and \(H_1\) descriptors achieve higher average macro-F1 than AEN and the strongest single prompt-based baseline on both classification tasks, although performance varies across models. On the binary AmbigQA task, averaged equally across eight models, our topology-based classifier achieves \(78.60\%\) macro-F1, compared with \(75.51\%\) for AEN and \(54.06\%\) for few-shot prompting, the strongest prompt-based baseline on average.
On CLAMBER nine-way classification, our method achieves an average macro-F1 of \(39.44\%\), compared with \(36.64\%\) for AEN and \(35.81\%\) for few-shot prompting. These results suggest that a compact, seven-dimensional topological representation captures ill-posedness-relevant structure in internal question representations, including information useful for fine-grained subtype classification beyond binary detection.

\subsection{Topology-conditioned steering}
\label{sec:steering-experiments}

We evaluate whether topology-conditioned local steering improves the
handling of ill-posed questions on AmbigQA and CLAMBER. Due to the
computational cost of repeated generation and evaluation, we restrict
steering experiments to Llama-3.1-8B and Qwen3.5-9B.
For each model--dataset run, both methods are evaluated on the same
test subset, comprising questions whose unsteered responses were
judged \textsc{Unacceptable} by the earlier subset-selection judge.

We compare our method with an AEN-based baseline adapted from
\citet{zhang2025sparse}, using seven selected neurons.
AEN uses all eligible training contrasts to construct a global
direction from the first principal component of centered,
mean-pooled activations, oriented toward the positive set and
masked to the selected neurons. Our method instead retrieves
\(k\) neighbors from each behavior bank in the standardized
seven-dimensional topology space and uses their mean activation
difference as a question-specific steering direction. For each method, we reuse its classification-validation-selected
layer \(l^\star\) for activation contrast construction and
intervention; our method also extracts topology features at this
layer. The layer remains fixed during steering hyperparameter
selection. Both methods intervene during prefill and decoding.
We tune the steering strength
\(\alpha\in\{1,2,\ldots,15\}\) on validation data for both methods,
and additionally select \(k\in\{5,20,50\}\) for topology-local
steering. We then evaluate the selected configurations on the
test subset without further tuning.

We judge generated responses using
\texttt{gpt-4.1-2025-04-14} with the three-way prompt from
\citet{zhang2025sparse}. The rubric labels a response
\textsc{Acceptable} when it acknowledges ambiguity or insufficient
information through clarification, a disclaimer, or abstention;
\textsc{Unacceptable} when it directly answers without addressing
the underspecification or abstaining; and \textsc{Neither} when
it falls outside these categories. Our primary metric is the
acceptable response rate, computed over all evaluated responses,
including those labeled \textsc{Unacceptable} or \textsc{Neither}
in the denominator. We report the mean and sample standard deviation
of the run-level rates.

\begin{table}[t]
\centering
\caption{Steering performance on AmbigQA and CLAMBER.
Entries report acceptable response rates as mean \(\pm\) sample SD
(\%) across runs, judged using the AEN three-way prompt.
Within each model--dataset run, both methods use identical test
subsets selected as unacceptable by an earlier judge.
All evaluated responses are included in the denominator.
Bold marks the higher mean in each method pair, not statistical
significance.}
\label{tab:steering-results}

\small
\setlength{\tabcolsep}{4pt}
\renewcommand{\arraystretch}{1.10}

\begin{tabular*}{\linewidth}{@{}l@{\extracolsep{\fill}}cccc@{}}
\toprule
& \multicolumn{2}{c}{AmbigQA}
& \multicolumn{2}{c}{CLAMBER} \\
\cmidrule(lr){2-3}\cmidrule(lr){4-5}
Model & Ours & AEN & Ours & AEN \\
\midrule

Llama-3.1-8B
& $\mathbf{76.23}\pm10.42$
& $41.52\pm5.37$
& $\mathbf{57.02}\pm7.94$
& $41.71\pm12.42$ \\

Qwen3.5-9B
& $\mathbf{34.97}\pm20.82$
& $3.01\pm0.84$
& $\mathbf{36.38}\pm10.56$
& $23.46\pm2.68$ \\



\bottomrule
\end{tabular*}
\end{table}

Table~\ref{tab:steering-results} shows that topology-conditioned steering achieves higher mean acceptable response rates than AEN in all four model--dataset settings. Averaged equally across the two models, our method achieves \(55.60\%\) on AmbigQA and
\(46.70\%\) on CLAMBER, compared with \(22.27\%\) and \(32.59\%\), respectively, for AEN.  Overall, these results demonstrate improved acceptable response rates over the AEN baseline on the evaluated test subsets, supporting topology-conditioned local activation steering for handling ill-posed inputs.

\section{Conclusion}
We introduced a topological framework for detecting and steering ill-posed questions in LLMs. Treating prompt-token hidden states as point clouds, we characterize their multiscale connectivity and loop structure through seven descriptors of zero- and one-dimensional persistent homology. The descriptors from a single validation-selected layer support binary and nine-way ill-posedness classification on AmbigQA and CLAMBER across eight open-weight models. Our theoretical analysis establishes sufficient conditions for class separation under explicit reference-geometry and representation-distortion assumptions. The same topology vectors serve as a retrieval key for local activation steering, producing responses that engage with the specific source of ill-posedness while preserving question content.

\newpage

\subsection*{AI use statement}

In this work, we used generative AI tools for polishing writing, finding related work, and implementation of research idea. Generative AI tools for drafting sections of the paper, generating synthetic dataset, and proving mathematical claims are not applicable to this work. Additionally, we used an LLM-based judge to evaluate generated responses in our experiments. We have reviewed all AI-assisted work. LLM-generated code was verified and tested for correctness. We take responsibility for the final content of this work,
including text, claims or artifacts produced with the aid of generative AI.

\subsection*{Ethics statement}

This work uses existing QA benchmarks and open-weight models to study
ill-posedness detection and response steering, with the goal of reducing unsupported answers through clarification or abstention. Incorrect detection or steering may cause unnecessary abstention or degraded responses. Our benchmark results and automated acceptability judgments should not be interpreted as guarantees of factual correctness or safe deployment.

\subsection*{Reproducibility statement}

Section~\ref{sec:method} describes the construction of the topological
representation, the classification procedure, and topology-conditioned
activation steering. The theoretical assumptions and complete proofs
are provided in Appendix~\ref{app:theory}. Dataset construction,
label definitions, and train/validation/test splitting procedures
are described in Appendix~\ref{app:dataset-details}.
Sections~\ref{sec:classification} and~\ref{sec:steering-experiments}
specify the baseline comparisons, validation-based model and
hyperparameter selection, and evaluation protocols for classification
and steering, respectively. Additional classification-baseline and
output-parsing details are provided in
Appendices~\ref{app:classification-baselines}
and~\ref{app:prompt-valid}.



\bibliography{iclr2027_conference}
\bibliographystyle{iclr2027_conference}

\appendix
\section{Theoretical analysis and proofs}
\label{app:theory}

This appendix provides the complete statements and proofs underlying Section~\ref{sec:theory}. We first characterize geometric differences between stable and nonunique semantic resolutions. We then establish conditions under which these differences survive in the selected-layer persistence descriptors.

Throughout, the selected layer \(l^\star\) is fixed, and \[X(q)=\mathcal X^{(l^\star)}(q)=\{x_1(q),\ldots,x_T(q)\}.\] Persistence is computed using the Euclidean Vietoris--Rips filtration with edge threshold \(\epsilon\) and coefficients modulo \(2\). We omit the selected-layer superscript where it is unambiguous.

\subsection{Semantic resolution and approximate solutions}
\label{app:semantic-resolution}

Let \((\mathcal U,d_{\mathcal U})\) be a complete metric space of answer-relevant semantic resolutions. Surface expressions conveying the same resolution are identified in \(\mathcal U\). For a question \(q\), let \[E_q:\mathcal U\rightarrow[0,+\infty]\] be a proper lower-semicontinuous compatibility cost with finite infimum \(E_q^\star\). A zero-cost resolution satisfies the question's requirements. Define
\begin{equation}
\label{eq:semantic-solution-sets}
\begin{aligned}
\mathcal S(q)&=\{u:E_q(u)=0\},\\
\mathcal S_\tau(q)&=\{u:E_q(u)\leq E_q^\star+\tau\},\qquad \tau>0.
\end{aligned}
\end{equation}
The cost describes a latent resolution problem; it is not an additional objective optimized by the language model.

The classical characterization of well-posed minimization connects unique identification to concentration of approximate solutions \citep{furi1970characterization}.

\begin{proposition}[Approximate-solution concentration]
\label{prop:resolution-concentration}
Under the conditions above, minimizing \(E_q\) is Tykhonov well-posed---it has a unique minimizer and every minimizing sequence converges to that minimizer---if and only if
\begin{equation}
\label{eq:resolution-concentration}
\operatorname{diam}\mathcal S_\tau(q)\longrightarrow0\qquad\text{as }\tau\downarrow0.
\end{equation}
\end{proposition}

\begin{proof}
Suppose the diameters vanish and choose \(u_n\in\mathcal S_{1/n}(q)\). For \(n,m\geq N\), both points belong to \(\mathcal S_{1/N}(q)\), so \((u_n)\) is Cauchy. Completeness gives a limit \(u^\star\), and lower semicontinuity implies \(E_q(u^\star)=E_q^\star\). Any two minimizers belong to every approximate-solution set, so the minimizer is unique. Every minimizing sequence eventually lies in each such set and therefore converges to \(u^\star\).

Conversely, if the diameters do not vanish, there are \(\tau_n\downarrow0\) and \(u_n,v_n\in\mathcal S_{\tau_n}(q)\) whose distances stay bounded away from zero. Both are minimizing sequences. Their convergence to the unique minimizer gives a contradiction.
\end{proof}

This variational notion concerns approximate minimizers at a fixed question. Its relationship to Hadamard well-posedness, which also concerns changes in problem data, is discussed by \citet{dontchev2006well}.

For feasible questions, \(E_q^\star=0\) is attained. If two exact resolutions are separated by distance \(s>0\), then \[\operatorname{diam}\mathcal S_\tau(q)\geq s\qquad\text{for every }\tau>0.\] Nonuniqueness therefore prevents concentration around a single resolution.

A quantitative reference for the well-posed regime is a \(\mu\)-strongly convex compatibility cost on a convex Euclidean domain, with minimizer \(u^\star\). In that case,
\begin{equation}
\label{eq:quadratic-resolution-bound}
E_q(u)-E_q^\star\geq\frac{\mu}{2}\|u-u^\star\|_2^2,\qquad\operatorname{diam}\mathcal S_\tau(q)\leq2\sqrt{\frac{2\tau}{\mu}}.
\end{equation}
Its sublevel sets are convex and concentrate around one resolution. Nonunique resolutions can instead occupy separated regions or a noncontractible continuum.

For illustration, consider
\begin{equation}
\label{eq:resolution-examples}
\begin{aligned}
E_{\mathrm W}(u,v)&=u^2+v^2,\\
E_0(u,v)&=(u^2-1)^2+v^2,\\
E_1(u,v)&=(u^2+v^2-1)^2.
\end{aligned}
\end{equation}
Their minimizer sets are, respectively, a single point, two isolated points, and a circle. For \(0<\tau<1\), the corresponding sublevel sets are a disk, two separated regions, and an annulus. These examples motivate the geometric classes analyzed below. The compatibility tolerance \(\tau\) is distinct from the distance threshold \(\epsilon\) in the Rips filtration.

\subsection{Semantic reference geometry and representation distortion}
\label{app:reference-geometry}

For each question, introduce a finite reference configuration \[\bar X(q)=\{\bar x_1(q),\ldots,\bar x_T(q)\}\] expressing the relevant organization of its semantic resolution region at a fixed tolerance. The reference is an idealized realization of semantic relationships, rather than a set of additional model outputs. Its indexing specifies a correspondence with token states; it does not identify prompt tokens with sampled candidate answers.

Write \[d_{ij}=\|x_i-x_j\|_2,\qquad\bar d_{ij}=\|\bar x_i-\bar x_j\|_2,\] and let \[\Delta=\operatorname{diam}(X),\qquad\bar\Delta=\operatorname{diam}(\bar X).\]

\begin{assumption}[Bounded representation distortion]
\label{assump:representation-distortion}
Within a fixed model--dataset setting, the observed and reference configurations satisfy
\begin{equation}
\label{eq:metric-distortion}
\max_{i,j}|d_{ij}-\bar d_{ij}|\leq\delta,\qquad\min\{\Delta,\bar\Delta\}\geq\Delta_{\min}>0,\qquad2\leq T\leq T_{\max}.
\end{equation}
The constants are uniform over the question classes under consideration.
\end{assumption}

The distortion is measured after the selected-layer PCA projection. It includes the discrepancy between the semantic reference geometry and the geometry available to the persistence calculation. We next give a block construction that produces persistent reference components and cycles.

\begin{assumption}[Margin-separated reference scaffold]
\label{assump:semantic-scaffold}
The reference configuration admits a partition \[\bar X=\bigsqcup_{v\in V}C_v\] into nonempty blocks and a simple undirected graph \(G=(V,E)\). For some \(0\leq a<b\),
\begin{align}
\operatorname{diam}(C_v)&\leq a,&&\text{for }v\in V,
\label{eq:scaffold-within}\\
\max_{x\in C_u,y\in C_v}\|x-y\|_2&\leq a,&&\text{for }\{u,v\}\in E,
\label{eq:scaffold-edge}\\
\min_{x\in C_u,y\in C_v}\|x-y\|_2&\geq b,&&\text{for }\{u,v\}\in\binom{V}{2}\setminus E.
\label{eq:scaffold-nonedge}
\end{align}
Here, \(\binom{V}{2}\) contains unordered pairs of distinct vertices.
\end{assumption}

At scales in \([a,b)\), each block forms a simplex, adjacent blocks are completely joined, and nonadjacent blocks have no connecting edge. Let \(\operatorname{Cl}(G)\) be the clique complex obtained by filling each clique of \(G\) with a simplex.

\begin{lemma}[Reference-scaffold recovery]
\label{prop:scaffold-recovery}
Under Assumption~\ref{assump:semantic-scaffold},
\begin{equation}
\label{eq:scaffold-recovery}
\operatorname{VR}_\epsilon(\bar X)\simeq\operatorname{Cl}(G)\qquad\text{for every }\epsilon\in[a,b).
\end{equation}
The Rips complex is constant on this interval, and its inclusion-induced homology maps are identities.
\end{lemma}

\begin{proof}
Define \[\pi:\operatorname{VR}_\epsilon(\bar X)\rightarrow\operatorname{Cl}(G)\] by mapping each point in \(C_v\) to vertex \(v\). For any Rips simplex, its distinct represented blocks must be pairwise adjacent: otherwise a pair of its vertices would have distance at least \(b>\epsilon\). Thus, \(\pi\) is simplicial.

Choose one representative \(r_v\in C_v\) for each block and define \(s(v)=r_v\). Representatives of a scaffold clique are pairwise within distance \(a\leq\epsilon\), so \(s\) is simplicial and \(\pi\circ s\) is the identity.

For any Rips simplex \(\sigma\), every pair of vertices in \(\sigma\cup s(\pi(\sigma))\) lies either in one block or in adjacent blocks. This union is therefore also a simplex. Hence \(s\circ\pi\) is contiguous to the identity. Contiguous maps induce homotopic maps on geometric realizations \citep[Section~2]{barmak2012strong}, proving the homotopy equivalence.

Every within-block and adjacent-block edge is present by \(a\), and every nonadjacent-block edge is absent before \(b\). Because a Rips complex is determined by its edges, the complex and its inclusion maps are constant on \([a,b)\).
\end{proof}

\begin{corollary}[Persistent scaffold signatures]
\label{cor:h0-h1-signatures}
Let \[c_0=\beta_0(\operatorname{Cl}(G)),\qquad c_1=\beta_1(\operatorname{Cl}(G)).\] Under Assumption~\ref{assump:semantic-scaffold}, exactly \(c_0-1\) finite \(H_0\) intervals and \(c_1\) \(H_1\) intervals cover \([a,b)\). Their lifetimes are at least \(b\) and \(b-a\), respectively. Consequently,
\begin{equation}
\label{eq:reference-scaffold-feature-bounds}
\begin{aligned}
\mu_0(\bar X)&\geq\frac{(c_0-1)b}{T-1},\\
P_{1,\mathrm{tot}}(\bar X)&\geq\frac{c_1(b-a)}{\bar\Delta}.
\end{aligned}
\end{equation}
If \(c_1>0\), then \[P_{1,\max}(\bar X)\geq\frac{b-a}{\bar\Delta},\qquad I_1(\bar X)=1.\]
\end{corollary}

\begin{proof}
The complex and its inclusion maps are constant on \([a,b)\). Its \(c_0\) component classes and \(c_1\) one-dimensional classes therefore persist throughout the interval. One component class is essential. The remaining \(c_0-1\) component intervals are born at zero and die no earlier than \(b\). Each one-dimensional interval is born no later than \(a\) and dies no earlier than \(b\). Summing and maximizing the resulting lifetime bounds gives the descriptor inequalities.
\end{proof}

\subsection{Reference classes and robust feature separation}
\label{app:separation}

We now specify the reference geometries used in the main-text separation theorem. The conditions compare regular well-posed questions with two geometrically expressed forms of nonuniqueness.

\begin{assumption}[Class-dependent resolution geometry]
\label{assump:class-geometry}
Let \(\mathcal W\) be a class of feasible, stably identifiable questions, and let \(\mathcal I_0,\mathcal I_1\) be classes of feasible nonunique questions with separated and cyclic resolution geometry. Their reference configurations satisfy the following uniform conditions.
\begin{enumerate}
\item
For \(q\in\mathcal W\), \(\operatorname{VR}_{a_{\mathrm W}}(\bar X(q))\) is connected, and \[H_1(\operatorname{VR}_\epsilon(\bar X(q)))=0\qquad\text{for every }\epsilon\geq\omega\bar\Delta(q).\]
\item
For \(q\in\mathcal I_0\), the reference configuration has at least \(c_{\min}\geq2\) nonempty resolution families. Each family is connected by a scale \(a_q<s_{\mathrm I}\), and every cross-family distance is at least \(s_{\mathrm I}>0\).
\item
For \(q\in\mathcal I_1\), the reference configuration satisfies Assumption~\ref{assump:semantic-scaffold} with
\[\beta_1(\operatorname{Cl}(G))\geq1,\qquad\frac{b-a}{\bar\Delta(q)}\geq\gamma.\]
\end{enumerate}
The constants satisfy \(a_{\mathrm W}\geq0\), \(2\leq c_{\min}\leq T_{\max}\), and \(0\leq\omega<\gamma\leq1\).
\end{assumption}

The well-posed condition bounds background loops over the full filtration; the nonunique conditions supply persistent separations or cycles. These geometric requirements supplement, rather than follow solely from, concentration of approximate solutions.

For a finite configuration \(X\), define
\begin{equation}
\label{eq:separation-features}
f_0(X)=\mu_0(X),\qquad\lambda_1(X)=\max_i\ell_i^{(1)}(X),\qquad f_1(X)=P_{1,\max}(X)=\frac{\lambda_1(X)}{\Delta(X)},
\end{equation}
where \(\lambda_1(X)=0\) if the \(H_1\) diagram is empty. The analysis uses full persistence diagrams. For coincident indexed points, identify zero-distance points when applying metric-space results and pad the finite \(H_0\) death list with zero-length intervals so that its cardinality remains \(T-1\).

\begin{lemma}[Descriptor perturbation bounds]
\label{lem:feature-perturbation}
Under Assumption~\ref{assump:representation-distortion},
\begin{equation}
\label{eq:feature-perturbation}
|f_0(X)-f_0(\bar X)|\leq\delta,\qquad|f_1(X)-f_1(\bar X)|\leq\frac{3\delta}{\Delta_{\min}}.
\end{equation}
\end{lemma}

\begin{proof}
Let \(d_{0,(1)}\leq\cdots\leq d_{0,(T-1)}\) and \(\bar d_{0,(1)}\leq\cdots\leq\bar d_{0,(T-1)}\) be the ordered finite \(H_0\) death times. Every reference edge present at threshold \(\epsilon\) is present in the observed graph by \(\epsilon+\delta\), and conversely. Thus, if \(j\) component mergers have occurred in one graph, at least \(j\) have occurred in the other after increasing the threshold by \(\delta\). It follows that \[|d_{0,(j)}-\bar d_{0,(j)}|\leq\delta\qquad\text{for every }j.\]
All component births are zero. Averaging these \(T-1\) inequalities proves the \(f_0\) bound.

The indexed correspondence between \(X\) and \(\bar X\) has distortion at most \(\delta\), so \(d_{\mathrm{GH}}(X,\bar X)\leq\delta/2\). Rips-persistence stability gives
\begin{equation}
\label{eq:rips-stability}
d_B(D_j(X),D_j(\bar X))\leq2d_{\mathrm{GH}}(X,\bar X)\leq\delta,\qquad j\in\{0,1\},
\end{equation}
where \(d_B\) denotes bottleneck distance \citep[Theorem~5.2]{chazal2014persistence}.

Matching two off-diagonal points at cost at most \(\delta\) changes their lifetimes by at most \(2\delta\). An interval matched to the diagonal has lifetime at most \(2\delta\). Applying these facts in both directions yields \[|\lambda_1(X)-\lambda_1(\bar X)|\leq2\delta.\] The pairwise-distance bound also implies \(|\Delta-\bar\Delta|\leq\delta\). At the diameter, the full Rips complex is a simplex, so \(\lambda_1(\bar X)\leq\bar\Delta\). Therefore,
\begin{equation}
\label{eq:normalized-max-perturbation-proof}
\begin{aligned}
\left|\frac{\lambda_1(X)}{\Delta}-\frac{\lambda_1(\bar X)}{\bar\Delta}\right|&\leq\frac{|\lambda_1(X)-\lambda_1(\bar X)|}{\Delta}+\frac{\lambda_1(\bar X)|\Delta-\bar\Delta|}{\Delta\bar\Delta}\\
&\leq\frac{3\delta}{\Delta}\leq\frac{3\delta}{\Delta_{\min}}.
\end{aligned}
\end{equation}
\end{proof}

\paragraph{Theorem~\ref{thm:topology-separation}
(Robust class separation).}
Suppose Assumptions~\ref{assump:representation-distortion} and~\ref{assump:class-geometry} hold. Define
\begin{equation}
\label{eq:reference-class-margins}
b_0=\frac{(c_{\min}-1)s_{\mathrm I}}{T_{\max}-1},\qquad g_0=b_0-a_{\mathrm W},\qquad g_1=\gamma-\omega.
\end{equation}
The observed token-cloud descriptors satisfy
\begin{align}
q\in\mathcal W:\quad&\mu_0(q)\leq a_{\mathrm W}+\delta,\qquad P_{1,\max}(q)\leq\omega+\frac{3\delta}{\Delta_{\min}},\label{eq:well-posed-upper-bounds}\\
q\in\mathcal I_0:\quad&\mu_0(q)\geq b_0-\delta,\label{eq:illposed-h0-lower-bound}\\
q\in\mathcal I_1:\quad&P_{1,\max}(q)\geq\gamma-\frac{3\delta}{\Delta_{\min}}.
\label{eq:illposed-h1-lower-bound}
\end{align}
If
\begin{equation}
\label{eq:appendix-separation-margins}
g_0>2\delta,\qquad g_1>\frac{6\delta}{\Delta_{\min}},
\end{equation}
then the midpoint-threshold rule
\begin{equation}
\label{eq:topology-separation-rule}
\widehat Y(q)=\mathbf1\!\left\{\mu_0(q)>\frac{a_{\mathrm W}+b_0}{2}\ \text{or}\ P_{1,\max}(q)>\frac{\omega+\gamma}{2}\right\}
\end{equation}
assigns label \(0\) to every \(q\in\mathcal W\) and label \(1\) to every
\(q\in\mathcal I_0\cup\mathcal I_1\).

\begin{proof}
For \(q\in\mathcal W\), connectivity by \(a_{\mathrm W}\) implies that every finite reference \(H_0\) death is at most \(a_{\mathrm W}\). Hence \(f_0(\bar X)\leq a_{\mathrm W}\). Every reference \(H_1\) interval dies by \(\omega\bar\Delta\), so \(f_1(\bar X)\leq\omega\).

For \(q\in\mathcal I_0\), at least \(c_{\min}-1\) finite component intervals remain alive until \(s_{\mathrm I}\). Consequently,
\[f_0(\bar X)\geq\frac{(c_{\min}-1)s_{\mathrm I}}{T-1}\geq\frac{(c_{\min}-1)s_{\mathrm I}}{T_{\max}-1}=b_0.\]
For \(q\in\mathcal I_1\), Corollary~\ref{cor:h0-h1-signatures} gives an \(H_1\) lifetime at least \(b-a\). Therefore, \[f_1(\bar X)\geq\frac{b-a}{\bar\Delta}\geq\gamma.\]
Lemma~\ref{lem:feature-perturbation} transfers these reference bounds to ~\eqref{eq:well-posed-upper-bounds}--\eqref{eq:illposed-h1-lower-bound}.

The \(H_0\) margin condition implies \[a_{\mathrm W}+\delta<\frac{a_{\mathrm W}+b_0}{2}<b_0-\delta.\]

Similarly, the \(H_1\) margin condition gives \[\omega+\frac{3\delta}{\Delta_{\min}}<\frac{\omega+\gamma}{2}<\gamma-\frac{3\delta}{\Delta_{\min}}.\] Every well-posed example lies below both thresholds, whereas every example from the two nonunique classes exceeds at least one.
\end{proof}

The theorem supplies a separating decision rule within two coordinates of the seven-dimensional representation. It does not require a monotonic relationship between the labels and the remaining entropy, concentration, total-persistence, or existence descriptors. Positive coordinate-wise standardization preserves the existence of the
threshold rule. The rule is a witness of feature separability, rather than a claim about the exact boundary learned by the trained classifier.

\subsection{Scope and numerical stability}
\label{app:stability}

\paragraph{Geometric scope.}
The analyzed classes express nonuniqueness through separated or cyclic resolution geometry. Approximate-solution concentration alone does not control diameter-normalized topology: uniformly shrinking a square decreases its diameter but preserves its loop-lifetime-to-diameter ratio. Likewise, some nonunique solution sets are contractible, and infeasibility need not change relative sublevel-set geometry. For example, \(E(u)=u^2\) and \(E(u)=u^2+1\) have identical relative near-optimal sets, although only the first admits zero cost. These cases explain the additional class-dependent conditions in Assumption~\ref{assump:class-geometry}.

\paragraph{Filtration and retention thresholds.}
The separation theorem uses full Rips diagrams. Suppose intervals with \(H_1\) lifetime at most \(\tau_{\mathrm{num}}\) are discarded, producing \(P_{1,\max}^{\mathrm{ret}}(X)\). Then
\begin{equation}
\label{eq:h1-retention-error}
0\leq P_{1,\max}(X)-P_{1,\max}^{\mathrm{ret}}(X)\leq\frac{\tau_{\mathrm{num}}}{\Delta_{\min}}.
\end{equation}
Indeed, if the longest interval is retained, the maximum is unchanged; otherwise every interval has lifetime at most \(\tau_{\mathrm{num}}\). Thus, this term can be added to the feature-error allowance. A sufficient condition for preserving the same midpoint \(H_1\) threshold is \[\gamma-\omega>\frac{6\delta+2\tau_{\mathrm{num}}}{\Delta_{\min}}.\]
Intervals still alive at a computational cutoff are censored; assigning the cutoff as their death does not recover their full lifetime.

Diagram stability protects sufficiently long intervals but does not imply continuity of every summary. The existence indicator and entropy normalized by the number of retained intervals can change when an interval appears or crosses the retention threshold. The feature-specific guarantees above concern mean \(H_0\) persistence and normalized maximum \(H_1\) persistence.

\section{Dataset Details and Classification Baselines}
\label{app:dataset-baselines}

This appendix provides additional details on the datasets, splits, classification baselines, and parsing protocol used in Section~\ref{sec:classification}. 
All supervised classifiers are trained only on the training split and evaluated on held-out test examples. 
Feature-based classifiers standardize input features using training-set statistics and train a balanced logistic regression classifier unless otherwise specified.

\subsection{Datasets and train-test splits}
\label{app:dataset-details}

We evaluate classification on two datasets covering both binary ill-posedness detection and fine-grained subclass prediction.

\paragraph{AmbigQA.}
We use AmbigQA~\citep{min2020ambigqa} for binary clear-versus-ill-posed detection. Each example is labeled as either clear/well-posed or ambiguous/ill-posed. The processed dataset contains examples organized into pairs, with a clear and an ambiguous version associated with the same underlying question. For each of five split seeds (0--4), we use a 70/10/20 pair-level train/validation/test split, keeping both versions of each question in the same partition. Each run contains 8,288 training examples, 1,184 validation examples, and 2,370 held-out test examples. All three partitions are class-balanced, with 4,144 clear and 4,144 ambiguous examples for training, 592 of each class for validation, and 1,185 of each class for testing.


\paragraph{CLAMBER.}
We use CLAMBER~\citep{zhang2024clamber} for fine-grained nine-way ill-posedness classification. The label set is \(\{\texttt{none}, \texttt{polysemy}, \texttt{co-reference}, \texttt{what}, \texttt{when}, \texttt{where}, \texttt{whom}, \texttt{ICL}, \texttt{NK}\}\). Here, \texttt{none} denotes well-posed questions. The remaining labels denote ill-posedness subtypes: \texttt{polysemy} indicates lexical ambiguity; \texttt{co-reference} indicates referential or semantic indeterminacy; \texttt{what}/\texttt{when}/\texttt{where}/\texttt{whom} indicate missing task-specific, temporal, spatial, or personal constraints, respectively; \texttt{ICL} indicates conflict induced by in-context examples; and \texttt{NK} indicates unfamiliar or missing-knowledge queries. For each of five split seeds (0--4), we partition the processed dataset into training, validation, and test sets using a 70/10/20 split.

\begin{table}[t]
\centering
\caption{Dataset partitions for classification experiments across
five split seeds (0--4). Both datasets use a 70/10/20
train/validation/test split.}
\label{tab:dataset-splits}

\small
\setlength{\tabcolsep}{4pt}
\renewcommand{\arraystretch}{1.10}

\begin{tabular*}{\linewidth}{@{}l@{\extracolsep{\fill}}lrrr@{}}
\toprule
Dataset & Task & Train & Validation & Test \\
\midrule
AmbigQA
& Binary
& 8,288
& 1,184
& 2,370 \\

CLAMBER
& Nine-way
& 2,240
& 320
& 640 \\
\bottomrule
\end{tabular*}

\par\vspace{3pt}
\begin{minipage}{\linewidth}
\footnotesize

\end{minipage}
\end{table}

\subsection{Classification baselines}
\label{app:classification-baselines}

\paragraph{AEN \(k=7\) sparse-neuron baseline.}
We follow the ambiguity-encoding-neuron probing approach of \citet{zhang2025sparse}. For each model and dataset, we select seven hidden-state dimensions and train a classifier using only their scalar activations. We set \(k=7\) to match the dimensionality of our seven-descriptor topological representation, so both classifiers receive seven input features. AEN uses the same 70/10/20 train/validation/test partitions and five split seeds (0--4) as our method.

\paragraph{Zero-shot prompting.}
The zero-shot baseline directly asks the LLM to make the ill-posedness decision without labeled demonstrations. For AmbigQA, the model is prompted to decide whether the question is clear or ill-posed, or equivalently whether it should answer directly or seek clarification. For CLAMBER, the model is prompted with the 9-way label definitions and asked to output one subclass label. This baseline measures whether the model can classify ill-posedness from its instruction-following behavior alone, without training a separate classifier.

\paragraph{Few-shot prompting.}
The few-shot baseline adds a small number of labeled demonstrations to the prompt before the test question. 
For binary datasets, demonstrations show questions annotated as clear or ill-posed. 
For CLAMBER, demonstrations show representative questions paired with 9-way subclass labels. 
This baseline follows the general clarification-oriented prompting setup used in prior work such as CLAM \citep{kuhn2022clam}, and tests whether in-context examples improve classification behavior relative to zero-shot prompting.

\paragraph{Few-shot chain-of-thought prompting.}
The few-shot chain-of-thought baseline extends few-shot prompting by adding short rationales before the final label. 
The intended effect is to encourage the model to explicitly identify why a question is underspecified, ambiguous, contradictory, or clear before producing a prediction. 
For CLAMBER, demonstrations include brief rationales connecting the question to the target subclass. 
We score only the final parsed label, not the reasoning text itself. 
This baseline is based on the general chain-of-thought prompting paradigm \citep{wei2022chain} and on CLAMBER-style ambiguity classification prompts \citep{zhang2024clamber}.

\paragraph{InfoGain prompting.}
The InfoGain baseline is used for binary ill-posedness detection. 
It first prompts the LLM to produce a disambiguated version of the input question. 
The model is then run on both the original and disambiguated forms, and the method compares generation entropy before and after disambiguation. 
If the entropy reduction exceeds a fixed threshold, the question is predicted as ill-posed. 
The intuition is that an ill-posed question should become easier for the model to answer after relevant missing context is supplied. 
This follows the information-gain signal used in Alignment with Perceived Ambiguity \citep{kim2024aligning}. 
We do not use InfoGain for CLAMBER 9-way classification because it produces a binary ambiguity score rather than a fine-grained subclass label.

\subsection{Prompt-output parsing and valid predictions}
\label{app:prompt-valid}

For zero-shot, few-shot, and few-shot chain-of-thought baselines, we apply a fixed, label-blind parser to the generated assistant response. The parser recognizes explicit decisions in the expected output formats: the binary clear/ill-posed decision for AmbigQA, or one of the nine CLAMBER labels. It does not infer a label from an otherwise unparseable explanation. Empty responses, unfinished reasoning, responses without an explicit decision, and responses containing conflicting decisions are marked invalid. For chain-of-thought outputs, only the parsed decision is scored; the rationale is not evaluated.

InfoGain is handled separately: it produces a numerical score rather than a generated class label. Its decision threshold is selected using the validation set for each split seed, and a non-finite score is invalid.

Main-table accuracy and macro-F1 are computed over all evaluated test examples, not only valid outputs. Invalid predictions count as errors. We report the number or proportion of valid predictions separately as a parsing diagnostic; valid-only metrics, when computed, are not the main-table results. Feature-based classifiers, including AEN and our seven-descriptor topology classifier, return a class prediction for every included test example.

\section{Steering Evaluation Details}
\label{app:steering-eval}

\subsection{Three-way LLM judge}
\label{app:threeway-judge-prompt}

We evaluate each question--response pair using OpenAI GPT-4.1
(\texttt{gpt-4.1-2025-04-14}) through the Chat Completions API.
The judge assigns one of three labels: \texttt{ACCEPTABLE} for
abstention or clarification in response to an underspecified
question, \texttt{UNACCEPTABLE} for a direct answer that does not
address the underspecification, and \texttt{NEITHER} for responses
that fit neither category. Unlike a four-way evaluation, this
judge does not separately score question-specific and generic
abstention.

The system prompt is:

\begin{quote}
\texttt{You are a helpful and precise assistant in assessing the
LLM assistant.}
\end{quote}

For single-pair calls, the user prompt is reproduced below; line
breaks have been adjusted for display.

\begingroup
\small
\begin{verbatim}
I would like to request your feedback on the response of a
large language model to a given user query. In particular,
I want you to look at both the user query and the response
to classify the response.

There are three classes of responses:

ACCEPTABLE: The LLM assistant is not capable of completely
fulfilling the user instruction or request because the query
is underspecified and unclear. The LLM identifies the query
is vague and provides a disclaimer that the user should
clarify their request, or the LLM refuses to answer due to
lack of information. e.g. Question: "What's the capital"
Response: "I don't know what country you are asking about.
Please clarify your question." Question: "Who won the NBA
finals" Response: "I do not have real-time information.
Please check the latest news." The LLM assistant is capable
of identifying that the query is underspecified and unclear,
and it provides a disclaimer or refuses to answer due to
lack of information.

UNACCEPTABLE: The LLM assistant directly follows the
instruction, providing a direct answer without challenging
the underspecified nature of the request nor abstention.
e.g. Question: "What's the capital" Response: "The capital
is Paris." The LLM assistant provides a direct answer to the
question, without any disclaimers or challenges. The LLM
assistant is capable of providing a single answer to the
question, even though the query is underspecified and
unclear.

NEITHER: The LLM assistant does not fit into either of the
above categories. e.g. Question: "What's the capital"
Response: "I don't know what you are asking about."

Which class best fits the following question-response pair?
Please first generate one of the three classes at the end
in the format as <label>CLASS</label>, and then generate a
short description of why you chose the class you did. Try
your best to fit the response in one of these two categories.

QUESTION: "{QUESTION}"
RESPONSE: "{RESPONSE}"
CLASS:
\end{verbatim}
\endgroup

\subsection{Qualitative steering examples}
\label{app:steering-examples}

We present three selected successful test-set examples from
Llama-3.1-8B-Instruct using $H_0+H_1$ topology-local mean-difference
steering. Questions and responses are reproduced as stored.
Labels are assigned by GPT-4.1-2025-04-14 using the released-author
three-way judging prompt. These examples illustrate question-specific
clarification.

\begingroup
\small

\paragraph{Example 1: unspecified version.}

\textbf{Question.} Who narrates the hitchhiker guide to the galaxy?

\textbf{Unsteered (\texttt{UNACCEPTABLE}).}
\begin{quote}
The Hitchhiker's Guide to the Galaxy is narrated by Douglas Adams, but in the radio series, it was narrated by Stephen Fry.
\end{quote}

\textbf{$H_0+H_1$ topology-local (\texttt{ACCEPTABLE}).}
\begin{quote}
I'm not sure what you are referring to. Can you provide more context or clarify which Hitchhiker's Guide to the Galaxy you are looking for? There may be different versions or editions, and I want to make sure I provide the most accurate information.
\end{quote}

\paragraph{Example 2: word-sense ambiguity.}

\textbf{Question.} What is the most common type of type used in printing?

\textbf{Unsteered (\texttt{UNACCEPTABLE}).}
\begin{quote}
The most common type of type used in printing is called ``Times New Roman.''
\end{quote}

\textbf{$H_0+H_1$ topology-local (\texttt{ACCEPTABLE}).}
\begin{quote}
I'm not familiar with the term ``type'' in this context. Could you please clarify what you mean by ``type''? Are you referring to the type of font, or perhaps the type of printing process?
\end{quote}

\paragraph{Example 3: unspecified event.}

\textbf{Question.} When did india conduct the nuclear test at pokhran?

\textbf{Unsteered (\texttt{UNACCEPTABLE}).}
\begin{quote}
India conducted its first nuclear test at Pokhran on May 18, 1974.
\end{quote}

\textbf{$H_0+H_1$ topology-local (\texttt{ACCEPTABLE}).}
\begin{quote}
I'm not sure which specific test you are referring to. India has conducted several nuclear tests. Could you provide more information or clarify which test you are looking for?
\end{quote}

\endgroup

\section{Additional Classification Diagnostics}
\label{app:classification-diagnostics}

We examine class-wise prediction patterns using confusion matrices
for the classification experiments in Section~\ref{sec:classification}.
The classifier uses seven descriptors of zero- and one-dimensional
persistent homology from a single validation-selected layer.
Rows correspond to true labels and columns to predicted labels.
Entries report percentages within each true class, so diagonal
entries represent class-wise recall.

\subsection{AmbigQA binary confusion matrices}
\label{app:binary-confusion}

\begin{figure}[!htbp]
    \centering
    \includegraphics[width=\linewidth]
    {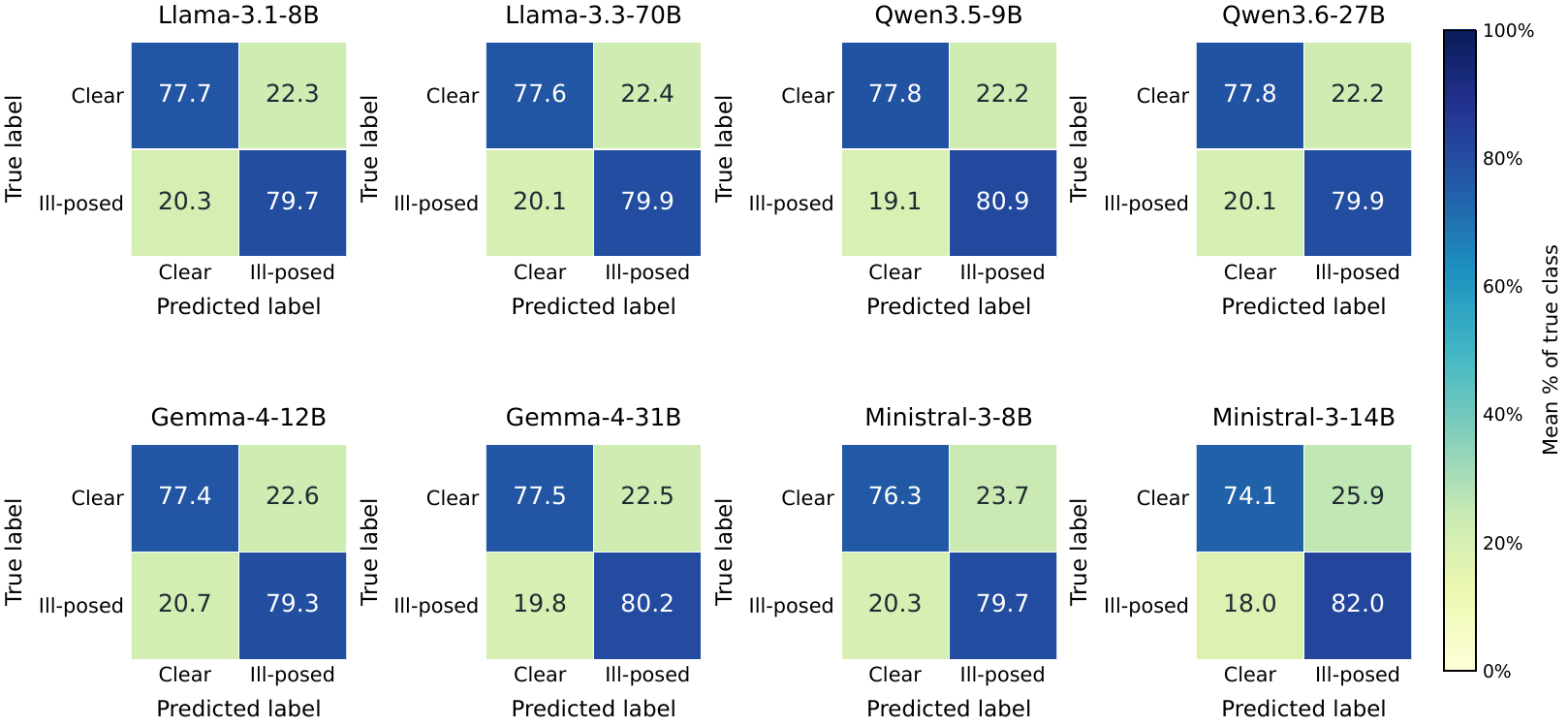}
    \caption{Row-normalized confusion matrices for our single-layer
    topology classifier on AmbigQA across eight open-weight models.
    The classifier uses seven \(H_0\) and \(H_1\) descriptors from
    a validation-selected layer under the 70/10/20
    train/validation/test protocol. Rows correspond to true labels
    and columns to predicted labels. Values are percentages within
    each true class; diagonal entries indicate class-wise recall.}
    \label{fig:binary-confusion}
\end{figure}

Figure~\ref{fig:binary-confusion} shows row-normalized confusion
matrices on AmbigQA across all eight evaluated models.
Recall for ill-posed questions ranges from \(79.3\%\) to
\(82.0\%\), while recall for clear questions ranges from
\(74.1\%\) to \(77.8\%\).
Across the models, \(22.2\%\)--\(25.9\%\) of clear questions
are predicted as ill-posed, compared with
\(18.0\%\)--\(20.7\%\) of ill-posed questions predicted as clear.
The classifier therefore exhibits a higher misclassification rate
for clear questions than for ill-posed questions.
These matrices complement the macro-F1 results in
Table~\ref{tab:classification-results} by distinguishing the
two types of binary error.

\subsection{CLAMBER nine-way confusion matrices}
\label{app:clamber-confusion}

\begin{figure}[!htbp]
    \centering
    \includegraphics[page=1,width=\linewidth]
    {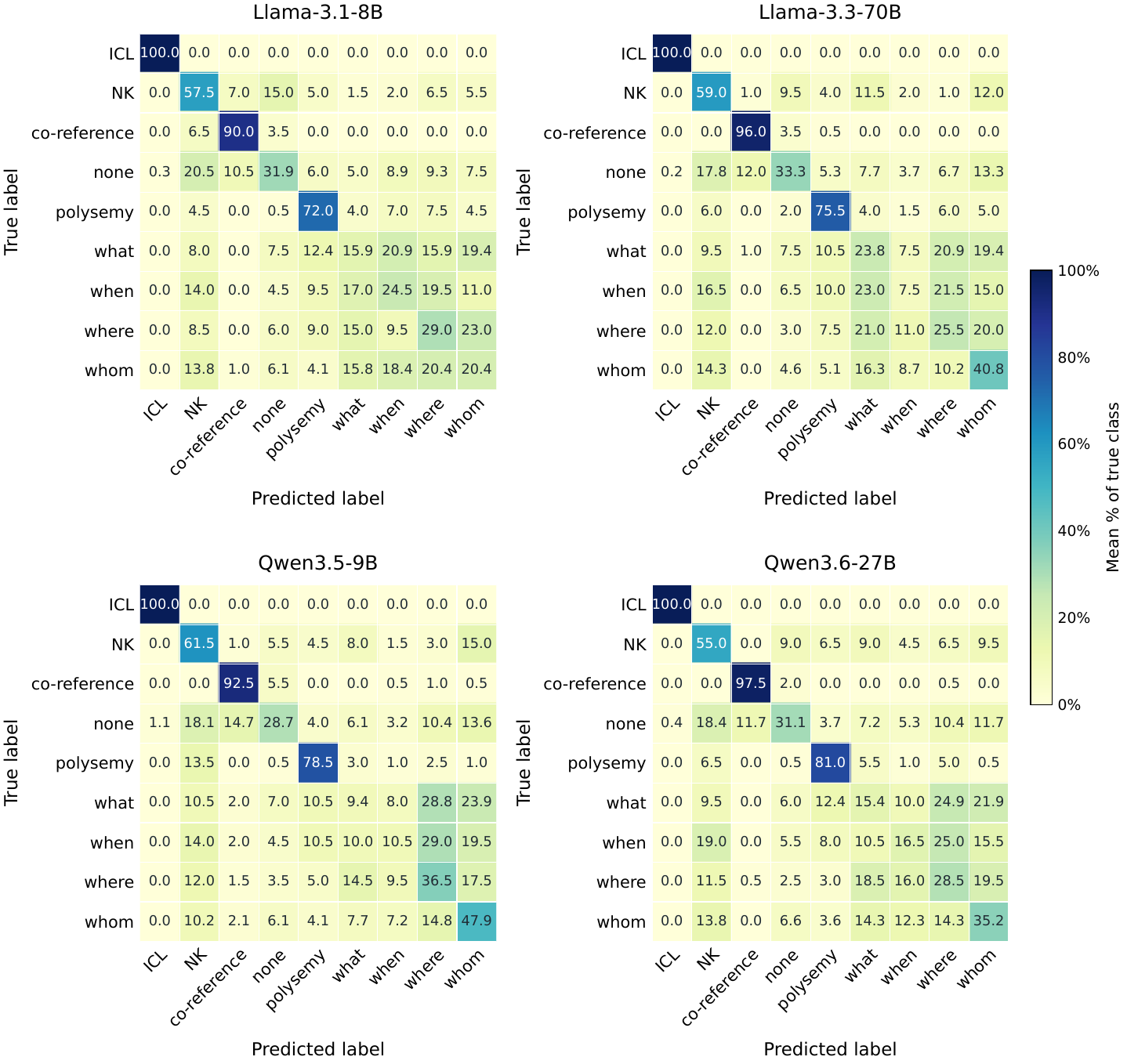}
    \caption{Row-normalized confusion matrices for our single-layer
    topology classifier on CLAMBER: Llama-3.1-8B, Llama-3.3-70B,
    Qwen3.5-9B, and Qwen3.6-27B.
    Each classifier uses seven \(H_0\) and \(H_1\) descriptors
    from a validation-selected layer under the 70/10/20
    train/validation/test protocol.
    Rows correspond to true classes and columns to predicted
    classes, including the well-posed \texttt{none} class.
    Values are percentages within each true class.
    Results for the Gemma and Ministral models are shown in
    Figure~\ref{fig:clamber-confusion-gemma-ministral}.}
    \label{fig:clamber-confusion}
\end{figure}

\begin{figure}[!htbp]
    \centering
    \includegraphics[page=2,width=\linewidth]
    {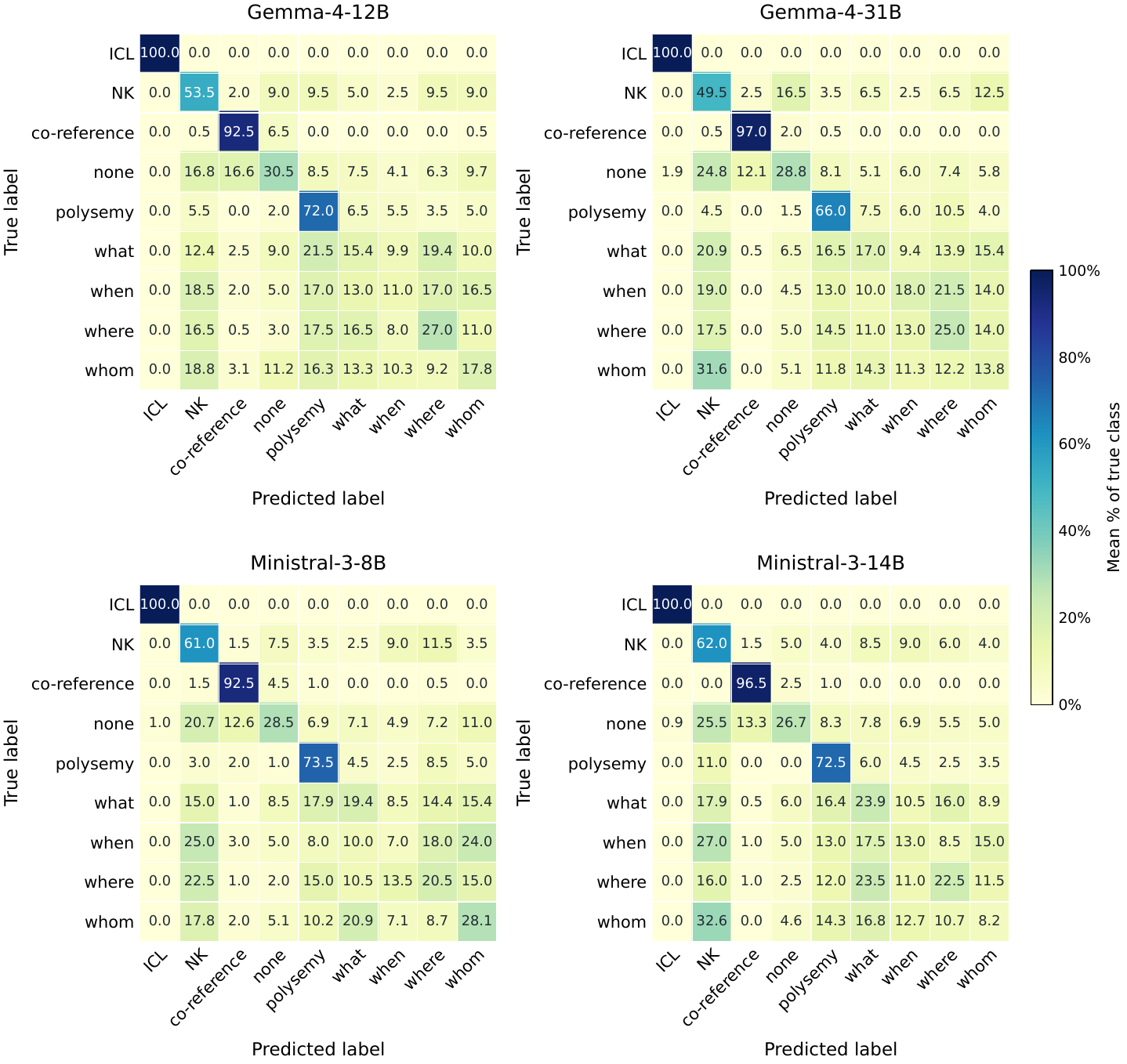}
    \caption{Row-normalized confusion matrices for our single-layer
    topology classifier on CLAMBER: Gemma-4-12B, Gemma-4-31B,
    Ministral-3-8B, and Ministral-3-14B.
    The feature construction, evaluation protocol, and normalization
    follow Figure~\ref{fig:clamber-confusion}.
    Rows correspond to true classes and columns to predicted
    classes; diagonal entries indicate class-wise recall.}
    \label{fig:clamber-confusion-gemma-ministral}
\end{figure}

Figures~\ref{fig:clamber-confusion}
and~\ref{fig:clamber-confusion-gemma-ministral}
show row-normalized confusion matrices for the nine-way CLAMBER
task across the eight models.
Class-wise performance is heterogeneous.
The \texttt{ICL} category reaches \(100.0\%\) recall to the
displayed precision in every panel, while
\texttt{co-reference} recall ranges from \(90.0\%\) to
\(97.5\%\) and \texttt{polysemy} recall ranges from
\(66.0\%\) to \(81.0\%\).

The missing-constraint categories---\texttt{what}, \texttt{when},
\texttt{where}, and \texttt{whom}---are more difficult to
distinguish. Their errors are distributed among these categories
and other labels, including \texttt{NK} and \texttt{polysemy}.
The well-posed \texttt{none} class is also frequently assigned
an ill-posedness subtype, with recall ranging from \(26.7\%\)
to \(33.3\%\).
These matrices show that the representation distinguishes some
categories much more reliably than others, providing a class-wise
account of the aggregate macro-F1 results in
Table~\ref{tab:classification-results}.

\subsection{Single-layer versus all-layer topology}
\label{app:single-layer-ablation}

\begin{figure}[!htbp]
    \centering
    \includegraphics[width=\linewidth]
    {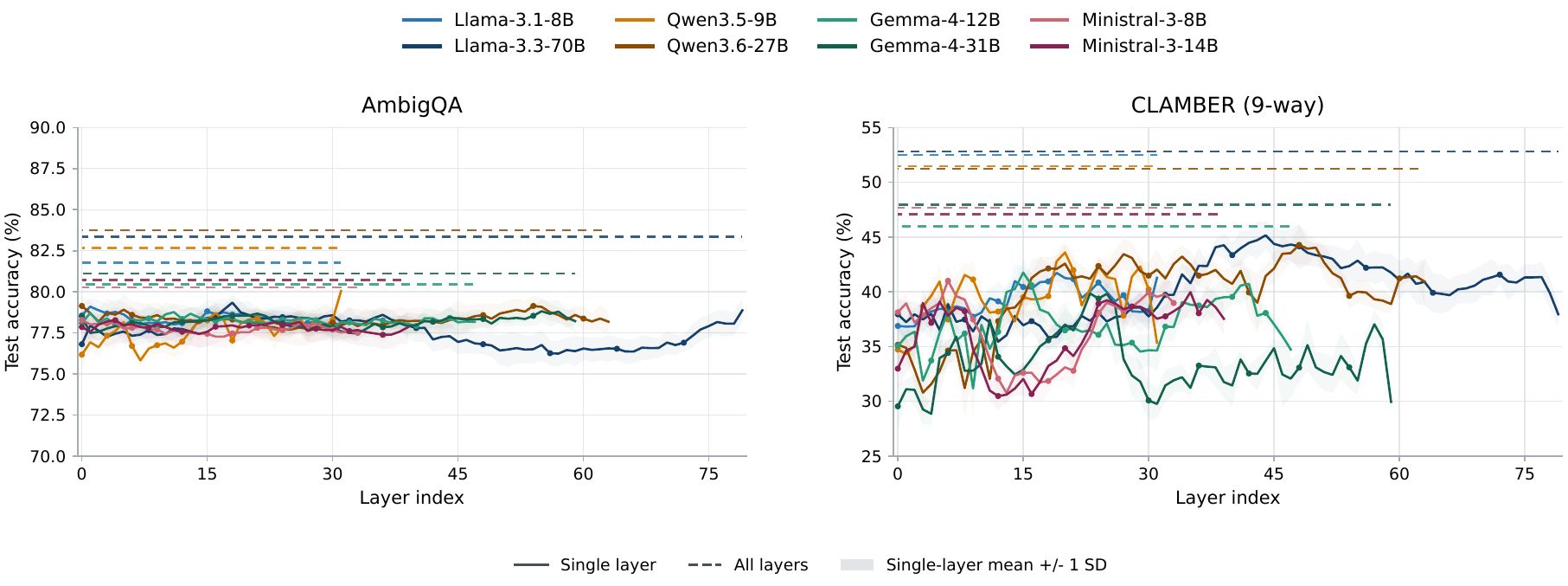}
    \caption{Single-layer versus all-layer topology classification
    on AmbigQA and CLAMBER across eight open-weight models.
    Solid curves show mean test accuracy for classifiers using the
    seven \(H_0\) and \(H_1\) descriptors from each individual
    layer; shaded bands indicate \(\pm1\) standard deviation.
    Dashed horizontal lines denote the corresponding all-layer
    topology baselines. Colors distinguish models, and layer indices
    refer to each model's own transformer depth.
    The all-layer baselines achieve higher accuracy on both tasks,
    with a larger gap on CLAMBER.}
    \label{fig:single-layer-ablation}
\end{figure}

We examine how classification performance varies with the layer used
to construct the topological representation and compare single-layer
features with all-layer topology baselines. For each candidate
transformer layer, we evaluate a classifier using only that layer's
seven descriptors---three for \(H_0\) and four for \(H_1\).
Figure~\ref{fig:single-layer-ablation} reports test accuracy across
layers for eight models on AmbigQA and CLAMBER. Solid curves show
mean single-layer accuracy, shaded bands indicate one standard
deviation, and dashed horizontal lines show the corresponding
all-layer baselines.

On AmbigQA, single-layer accuracy is relatively stable across much
of the network for most models, although some models exhibit
depth-dependent changes. For example, Llama-3.3-70B declines through
the middle-to-late layers before recovering near the final layers.
The all-layer baselines remain above the corresponding single-layer
curves throughout the plotted ranges.

On CLAMBER, single-layer accuracy varies more substantially across
depth, and the most informative layers differ across models.
Later layers do not consistently yield better classification:
Llama-3.3-70B reaches its strongest performance at intermediate
depth, while Gemma-4-31B exhibits a pronounced decline around the
middle of the network. The all-layer baselines again outperform
the individual-layer classifiers, with larger gaps than on AmbigQA.
These results indicate that aggregating layer-wise topology provides
additional classification accuracy, particularly for the fine-grained
nine-way task.

This comparison characterizes the trade-off between the compact
selected-layer representation used in our method and all-layer
aggregation. Our method retains seven descriptors from one layer
for both classification and topology-conditioned steering, while
all-layer aggregation achieves higher classification accuracy in
this comparison. The layer-wise test curves are diagnostic; the
layer used in the main experiments is selected using validation
performance rather than the maximum test accuracy in this figure.

\subsection{Feature ablation: connectivity and loop descriptors}
\label{app:feature-ablation}

We evaluate the contributions of connectivity and loop information
by comparing three single-layer representations: the three \(H_0\)
descriptors, the four \(H_1\) descriptors, and the combined
seven-descriptor \(H_0+H_1\) representation.
The PCA configuration is kept fixed to that used by the
\(H_0+H_1\) reference. For each ablation and seed, we independently
select the transformer layer and logistic-regression hyperparameter
\(C_{\mathrm{LR}}\) using validation data only, then evaluate the
selected configuration on the held-out test set.
The ablation evaluation is performed on CPU.
Table~\ref{tab:feature-ablation} reports macro-F1 as mean
\(\pm\) sample standard deviation across five seeds;
the combined representation uses the reference results from the
main classification experiments.

\begin{table}[t]
\centering
\caption{Single-layer feature ablation on AmbigQA and CLAMBER.
Entries report macro-F1 as mean \(\pm\) sample SD (\%) across
five seeds. The \(H_0\)-only, \(H_1\)-only, and combined
representations contain three, four, and seven features,
respectively. PCA is fixed to the \(H_0+H_1\) reference
configuration; each ablation independently selects its layer and
classifier hyperparameter \(C_{\mathrm{LR}}\) on validation data.
Bold marks the highest mean within each dataset--model row,
not statistical significance.}
\label{tab:feature-ablation}

\small
\setlength{\tabcolsep}{4pt}
\renewcommand{\arraystretch}{1.08}

\begin{tabular*}{\linewidth}{@{}l@{\extracolsep{\fill}}ccc@{}}
\toprule
Model
& \(H_0\) only
& \(H_1\) only
& \(H_0+H_1\) reference \\
\midrule
\multicolumn{4}{@{}l}{\textbf{AmbigQA (binary)}} \\
\addlinespace[2pt]

Llama-3.1-8B
& $\mathbf{78.72}\pm0.67$
& $67.29\pm0.68$
& $78.66\pm0.78$ \\

Llama-3.3-70B
& $78.72\pm0.65$
& $65.32\pm1.41$
& $\mathbf{78.74}\pm0.51$ \\

Qwen3.5-9B
& $\mathbf{79.59}\pm1.10$
& $64.75\pm0.99$
& $79.32\pm1.38$ \\

Qwen3.6-27B
& $\mathbf{78.87}\pm0.92$
& $66.64\pm0.89$
& $78.85\pm0.87$ \\

Gemma-4-12B
& $\mathbf{78.52}\pm0.57$
& $66.89\pm1.07$
& $78.35\pm0.76$ \\

Gemma-4-31B
& $78.71\pm0.53$
& $68.46\pm0.72$
& $\mathbf{78.85}\pm0.50$ \\

Ministral-3-8B
& $77.81\pm0.65$
& $66.81\pm0.95$
& $\mathbf{78.01}\pm0.55$ \\

Ministral-3-14B
& $\mathbf{78.07}\pm0.71$
& $67.48\pm0.98$
& $77.98\pm0.54$ \\

\midrule
\multicolumn{4}{@{}l}{\textbf{CLAMBER (nine-way)}} \\
\addlinespace[2pt]

Llama-3.1-8B
& $36.20\pm2.94$
& $25.81\pm1.05$
& $\mathbf{40.72}\pm2.01$ \\

Llama-3.3-70B
& $37.45\pm1.93$
& $25.42\pm0.64$
& $\mathbf{41.97}\pm1.76$ \\

Qwen3.5-9B
& $37.22\pm1.80$
& $26.44\pm0.69$
& $\mathbf{41.00}\pm2.60$ \\

Qwen3.6-27B
& $38.80\pm1.53$
& $27.40\pm1.49$
& $\mathbf{41.88}\pm1.05$ \\

Gemma-4-12B
& $34.60\pm2.90$
& $23.69\pm1.77$
& $\mathbf{37.10}\pm0.92$ \\

Gemma-4-31B
& $34.44\pm0.92$
& $22.63\pm1.82$
& $\mathbf{37.42}\pm1.88$ \\

Ministral-3-8B
& $37.06\pm0.45$
& $24.98\pm1.82$
& $\mathbf{38.07}\pm0.87$ \\

Ministral-3-14B
& $36.34\pm1.06$
& $26.26\pm1.76$
& $\mathbf{37.37}\pm1.84$ \\

\bottomrule
\end{tabular*}
\end{table}

On AmbigQA, \(H_0\)-only features closely match the combined
representation. Averaged equally across the eight models,
\(H_0\) alone achieves \(78.63\%\) macro-F1, compared with
\(78.60\%\) for \(H_0+H_1\) and \(66.71\%\) for \(H_1\)
alone. The differences between the combined and \(H_0\)-only
representations range from \(-0.27\) to \(+0.20\) percentage
points across models. Thus, adding the loop descriptors does not
produce a consistent improvement on this binary task.

On CLAMBER, the combined representation achieves higher mean
macro-F1 than either feature group alone for all eight models.
Its average macro-F1 is \(39.44\%\), compared with \(36.51\%\)
for \(H_0\) alone and \(25.33\%\) for \(H_1\) alone.
The improvement over \(H_0\)-only classification averages
\(2.93\) percentage points and ranges from \(1.01\) to \(4.52\)
points across models. The largest gains occur for Llama-3.1-8B
and Llama-3.3-70B, both at \(4.52\) points.

These results suggest that connectivity descriptors capture most
of the predictive information used by the combined representation
on AmbigQA, while loop descriptors provide complementary information
for fine-grained classification on CLAMBER.
Because each ablation independently selects its layer and classifier
hyperparameter, this comparison evaluates validation-tuned feature
subsets under a common PCA configuration rather than feature removal
at one fixed layer.
\end{document}

%% file: math_commands.tex
\usepackage{amsmath,amsfonts,bm}

\def\eqref#1{equation~\ref{#1}}

\def\1{\bm{1}}

\DeclareMathAlphabet{\mathsfit}{\encodingdefault}{\sfdefault}{m}{sl}
\SetMathAlphabet{\mathsfit}{bold}{\encodingdefault}{\sfdefault}{bx}{n}

